\documentclass{article}
\pdfoutput=1
\usepackage{arxiv}

\usepackage[utf8]{inputenc} 
\usepackage[T1]{fontenc}    
\usepackage{hyperref}       
\usepackage{url}            
\usepackage{booktabs}       
\usepackage{amsfonts}       
\usepackage{nicefrac}       
\usepackage{microtype}      

\usepackage{lipsum}         
\usepackage{graphicx}
\usepackage[square,sort,comma,numbers]{natbib}
\usepackage{doi}
\usepackage{amsmath}
\usepackage{cleveref}       

\usepackage[normalem]{ulem}
\usepackage[ruled]{algorithm2e}
\SetAlCapSkip{1ex}
\SetKwInput{KwInput}{Input}
\SetKwInput{KwOutput}{Output}
\SetKwInput{KwParameter}{Parameter}
\SetKwInput{KwParameters}{Parameters}

\usepackage{authblk}

\author[1]{Li-Wen Chang$^*$}
\author[1]{Wenlei Bao$^*$}
\author[1]{Qi Hou$^*$}
\author[1]{Chengquan Jiang$^*$}
\author[1]{Ningxin Zheng$^*$}
\author[2]{Yinmin Zhong$^*$}
\author[1]{\protect\\Xuanrun Zhang$^*$}
\author[1]{Zuquan Song}
\author[1]{Ziheng Jiang}
\author[1]{Chengji Yao}
\author[1]{Haibin Lin}
\author[2]{Xin Jin}
\author[1]{Xin Liu}
\affil[1]{ByteDance Ltd}

\affil[ ]{\textit {\{liwen.chang, wenlei.bao, houqi.1993, jiangchengquan, zhengningxin,\protect\\zhangxuanrun, zuquan.song, chengji.yao, ziheng.jiang, haibin.lin, liuxin.ai\}@bytedance.com}}

\affil[2]{Peking University}
\affil[ ]{\textit {\{zhongyinmin, xinjinpku\}@pku.edu.cn}}

\begin{document}
\def\flux{Flux\xspace} 
\newcommand{\ignore}[1]{} 

\title{\flux: Fast Software-based Communication Overlap on GPUs through Kernel Fusion}
\maketitle
\def\thefootnote{*}\footnotetext{These authors contributed equally to this work}\def\thefootnote{\arabic{footnote}}
\begin{abstract}
Large deep learning models have demonstrated strong ability to solve many tasks across a wide range of applications. 
Those large models typically require training and inference to be distributed. 
Tensor parallelism is a common technique partitioning computation of an operation or layer across devices to overcome the memory capacity limitation of a single processor, and/or to accelerate computation to meet a certain latency requirement. 
However, this kind of parallelism introduces additional communication that might contribute a significant portion of overall runtime. 
Thus limits scalability of this technique within a group of devices with high speed interconnects, such as GPUs with NVLinks in a node.

This paper proposes a novel method, \flux, to significantly hide communication latencies with dependent computations for GPUs. 
\flux overdecomposes communication and computation operations into much finer-grained operations and further fuses them into a larger kernel to effectively hide communication without compromising kernel efficiency.  
\flux can potentially overlap up to 96\% of communication given a fused kernel. 
Overall, it can achieve up to 1.24x speedups for training over Megatron-LM on a cluster of 128 GPUs with various GPU generations and interconnects, and up to 1.66x and 1.30x speedups for prefill and decoding inference over vLLM on a cluster with 8 GPUs with various GPU generations and interconnects.
\end{abstract}
\pagestyle{plain} 

\thispagestyle{empty}

\section{Introduction}
\label{sec:info}

In the rapidly evolving field of deep learning, one of the most significant trends recently has been the development of increasingly large models~\cite{kaplan2020scaling}. 
This progression towards large models is not merely a pursuit of scale for its own sake, but a strategic response to the diverse and complex challenges encountered across various domains. These large models have demonstrated remarkable proficiency in tasks ranging from natural language processing~\cite{brown2020language, chowdhery2022palm, smith2022using}, computer vision~\cite{ramesh2021zeroshot, dehghani2023scaling}, to speech recognition~\cite{radford2022robust, Zhang_2022}, showcasing their versatility and effectiveness. By leveraging vast amounts of data and computational power, they have been able to unearth intricate patterns and insights that were previously inaccessible, offering unprecedented opportunities in fields as varied as healthcare~\cite{liu2023large}, finance~\cite{wu2023bloomberggpt}, software development~\cite{chen2021evaluating}, and beyond. 
This growth in model size correlates strongly with enhanced performance, opening new frontiers in artificial intelligence applications and redefining what machines are capable of achieving.

\begin{figure}[t]
\centering
\includegraphics[scale=0.5, bb=0 0 830 450]{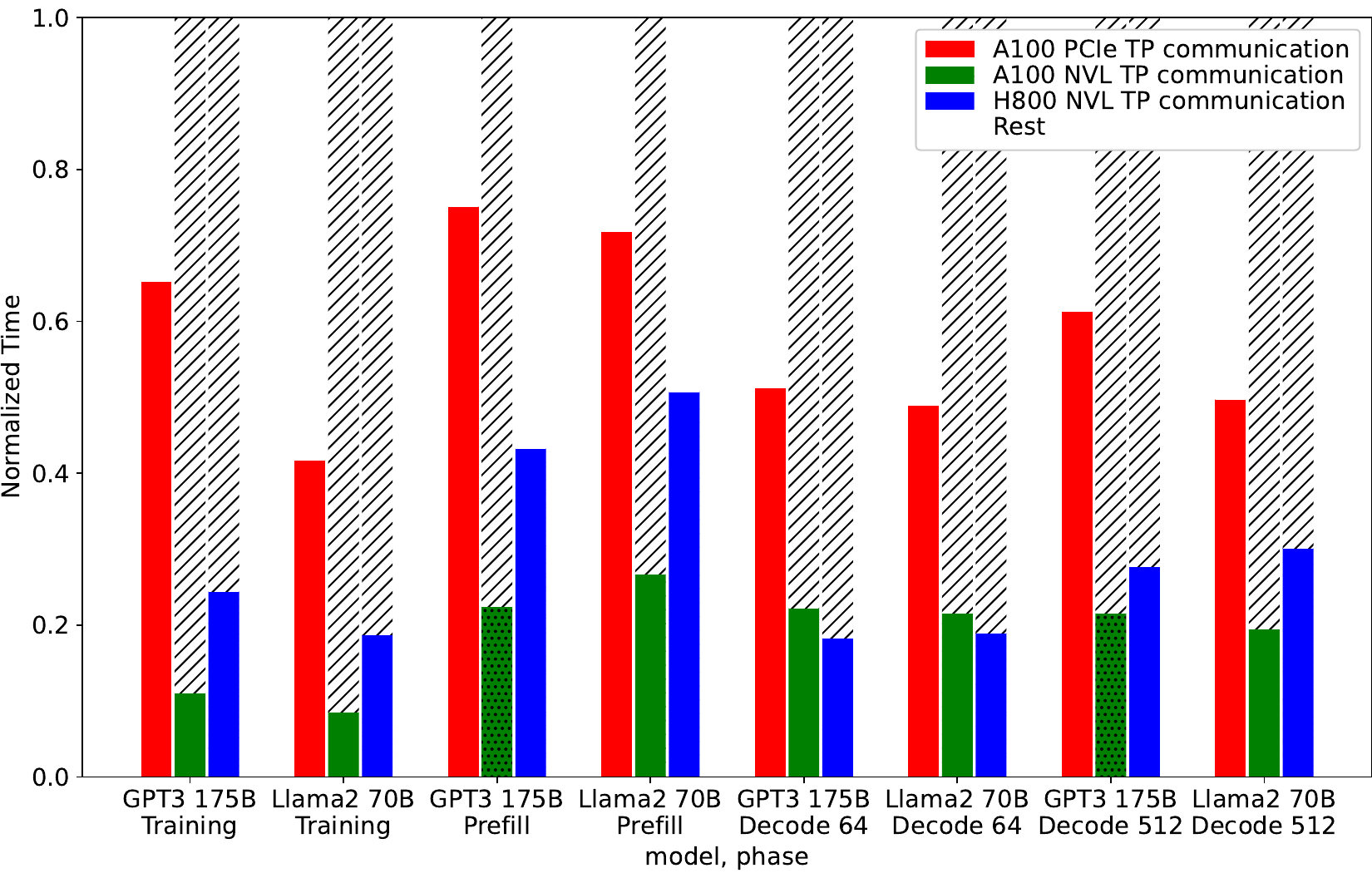}
\caption{Non-overlapped communication portion within tensor parallelism in common LLM workloads for training with 2-way data, 8-way pipeline, 8-way tensor parallelism on various 128-GPU clusters, and inference with 8-way tensor parallelism on various 8-GPU clusters.}
\label{fig:comm_time}
\end{figure}

These large deep learning models typically require training and inference to be distributed, due to their parameters well beyond the memory capacity of one single device. 
Model parallelism, such as tensor and pipeline parallelism, is applied to overcome this limitation. 
While pipeline parallelism partitions a model across layers into multiple devices, executing multiple batches in a pipeline fashion, 
tensor parallelism partitions an individual layer into multiple devices, executing in parallel. 
Both are important and could be applied together, but they do have different characteristics.  
Compared to pipeline parallelism improving throughput, tensor parallelism can shrink latency,
which is critical for inference.
Since tensor parallelism partitions a layer into multiple devices, additional data communication across devices might be required for gathering or (re-)distributing correct data, especially when a consecutive layer applies a different partitioning strategy or consumes data across partitions.
Figure~\ref{fig:comm_time} shows the substantial portion of communication time over the overall runtime for training and inference specifically for applying tensor parallelism, demonstrating the motivation and strong need to reduce the exposed communication time. 

Communication overlapping techniques~\cite{jangda2022breaking, wang2023overlap, te, lamy2023breadth, narayanan2021efficient, li2023fold3d, shah2023taccl, cowan2023mscclang, wang2023zero++} have become crucial for various kinds of parallelism in training and inferring large deep learning models. The existing overlapping techniques~\cite{jangda2022breaking, wang2023overlap, te} for tensor parallelism decompose a communication operation along with the dependent computation operation into a sequence of chunk, point-to-point operations based on the number of partitions,
and carefully execute paired decomposed communication and computation with no data dependence in parallel. 
These methods might have several limitations on GPUs, such as no precise control of execution timing on GPUs when using streams, and poor GPU utilization for executing multiple smaller kernel instances. 

To better overlap communication without compromising GPU utilization, we propose a new overlapping method, Fine-grained Communication Overlapping (\flux), that decomposes the original communication and computation into much finer-grained tiles than the existing methods, and then fuses tiled computation and communication into a single larger kernel. 
In the fused kernel, each dependent computation and communication tile is mapped into each thread block\footnote{Depending on GEMM implementations, a tile can also be  mapped into a warp or a thread block cluster as well.}. 
\flux optimizes communication together with computation, including kernel fusion, tile coordinate swizzling, GPU instruction selection, communication order selection, etc., and \flux could better adapt the GPU architecture as well as the interconnect over the existing methods. 
On top of that, \flux is built in a modular way by adopting NVIDIA CUTLASS~\cite{cutlass}, and can be easily auto-tuned across various combinations of GPU architectures and interconnects.
Therefore, \flux can deliver more efficient communication overlapping over the existing methods. 

Overall, we make the following contributions in this paper: 
\begin{itemize}
  \item We identify several performance issues when applying the existing communication overlapping techniques for tensor parallelism on GPUs.
  
  \item We propose a new novel communication overlapping technique that overcomes the above issues and naturally fits the modern GPU design. 

  \item We implement the proposed technique using NVIDIA CUTLASS with multiple optimizations for various GPU generations (A100 and H800), and various intra-node interconnects (PCIe and NVLink). 
  
  \item  We evaluate the proposed technique on various 128-GPU clusters for training and 8-GPU clusters for inference. The evaluation results demonstrate up to 1.24x, 1.66x, and 1.30x speedups over the non-overlapping method, such as Megatron-LM~\cite{megatron} and vLLM~\cite{kwon2023efficient}, and 1.38x, 2.06x, and 2.10x speedups over the previous overlapping method, TransformerEngine~\cite{te}, for training, prefill, and decoding, respectively.
\end{itemize}

\section{Backgrounds on Tensor Parallelism and Overlapping}
\label{sec:background}

Tensor parallelism, also called intra-layer model parallelism, is a technique partitioning a layer of a model over multiple devices. 
This section covers the common state-of-the-art partitioning patterns widely used in large deep learning models, and the corresponding conventional communication overlap strategies. 

\subsection{Common Partitioning and Communication Patterns}

\begin{figure}[t]
\centering
\includegraphics[scale=0.8, bb=0 0 200 350]{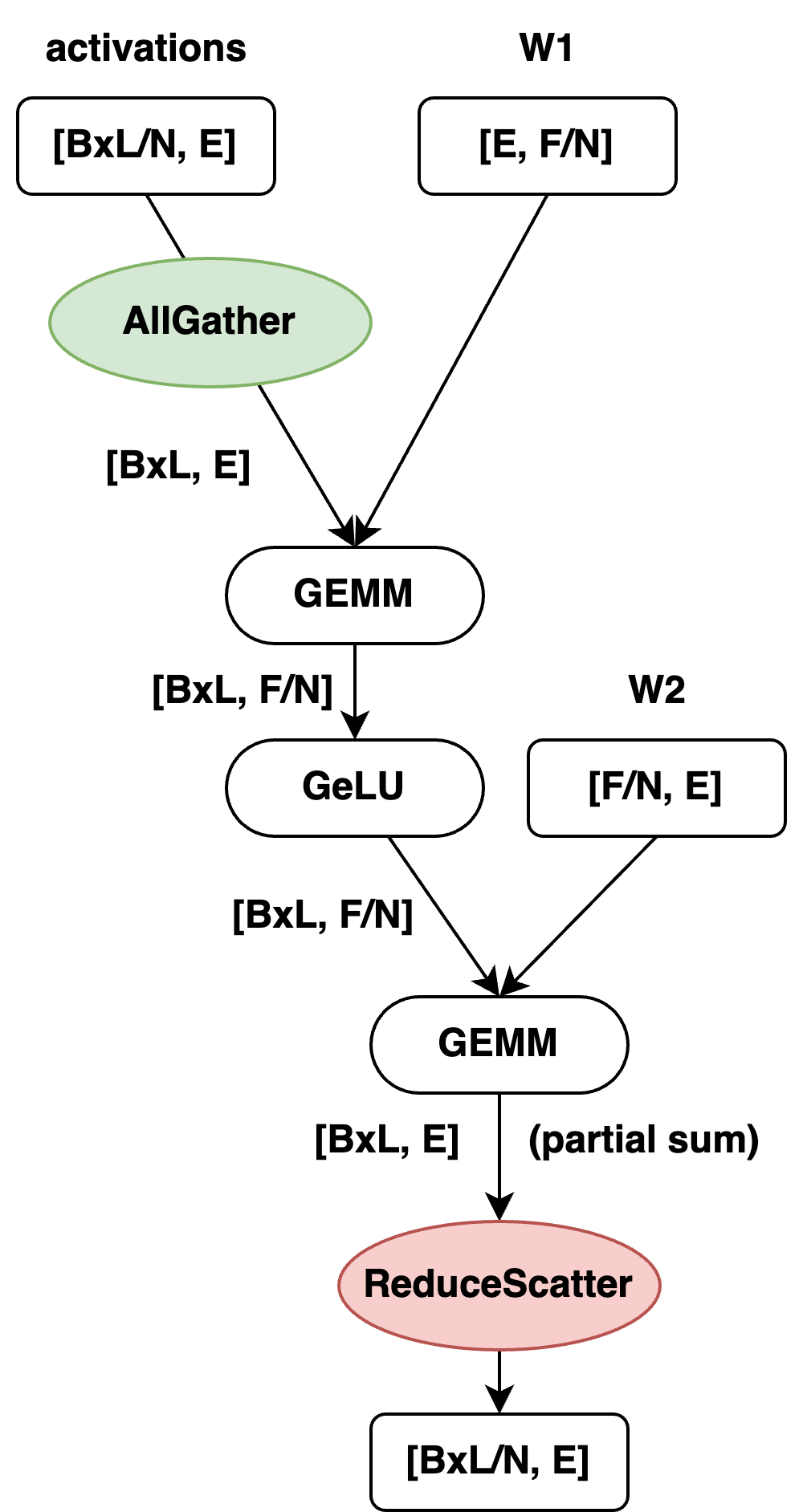}
\caption{Forward-propagation of the MLP portion with a \textit{N}-way partitioning across N devices.  
Here, B and L are flattened to fit the common notation of GEMM. 
}
\label{fig:tp}
\end{figure}

The common partitioning strategy we discuss in the paper is an extended Megatron-LM~\cite{shoeybi2019megatron} with sharded activation~\cite{korthikanti2023reducing, wang2023overlap}. 
For the sake of brevity, we use a multi-layer perception (MLP) portion within a transformer as an example to explain common partitioning and communication patterns.
Strategies for other operations, such as multi-head attention, or multi-query attention, can be found in  
~\cite{shoeybi2019megatron, korthikanti2023reducing, pope2023efficiently}. 

Figure~\ref{fig:tp} shows the common communication partitioning pattern in forward-propagation of the MLP example. 
The first GEMM shards the weight (W1) along the the row direction, and \textit{AllGather}s
the sharded input activations along the column direction before GEMM, 
while the second GEMM shards the weight (W2) along the column direction, and \textit{ReduceScatter}s the output activation along the column direction. 
In backward-propagation, \textit{AllGather}s and \textit{ReduceScatter}s are interchanged. 
As the figure shown, the dimensions of these two GEMM operations depends on the degree of tensor parallelism ($N$).
To avoid confusion, in the later sections, when describing problem sizes,   we would use a global, original shape, such as $[E, F]$, instead of a local shapes $[E, F/N]$, unless specified otherwise. 

Another common partitioning pattern that further shards all weights (W1 and W2) over devices involved in data parallelism, and \textit{AllGather}s weights before GEMM~\cite{rajbhandari2020zero, ren2021zero, rajbhandari2021zero, zhao2023pytorch, wang2023zero++}. 
In this partitioning pattern, 
since all weights have no data dependence before its consumer GEMM operations, those AllGather operations can be easily prefetched and overlapped with independent operations. 
Therefore, in the paper, we mainly discuss the first pattern.

\begin{figure}[t]
\centering
\includegraphics[scale=0.5,bb=0 0 850 450]{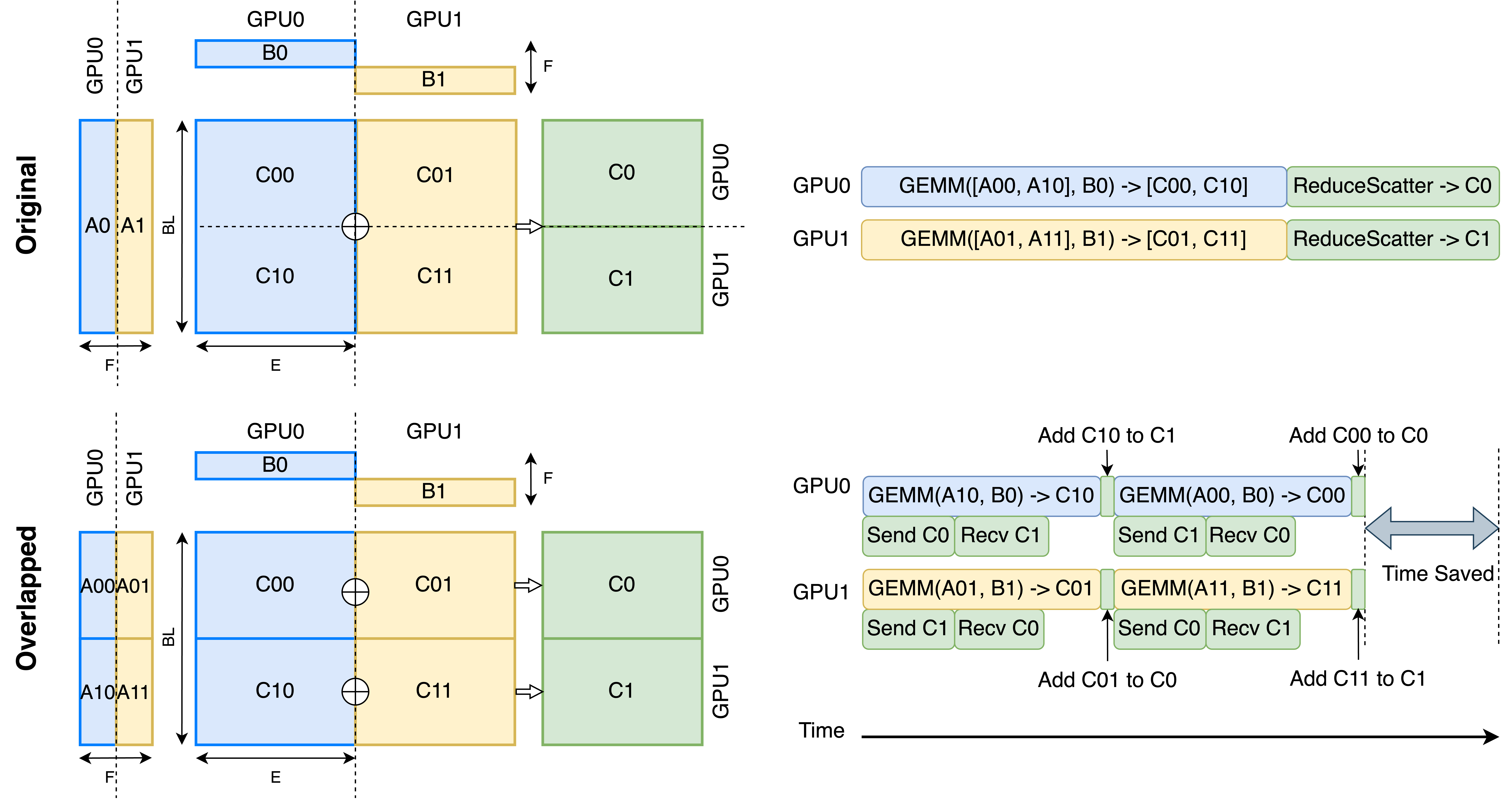}
\caption{An illustration of the prior GEMM-ReduceScatter overlapping with 2-way tensor parallelism.}
\label{fig:existing_rs}
\end{figure}

\subsection{Conventional Communication Overlapping Strategies}
Conventional methods~\cite{jangda2022breaking, wang2023overlap, te, jiang2024megascale} decompose the original computation and communication operations into chunks.
Then, carefully scheduling operations potentially can overlap communication with computation. 
The number of partitions in decomposition is aligned with the number of devices in tensor parallelism (or twofold of it to better utilize bidirectional data transfer). 
Limiting the number of partitions can potentially avoid complicating scheduling and reduce possible scheduling overheads. 
Figure~\ref{fig:existing_rs} illustrates a ReduceScatter overlapping scenario\footnote{Note the existing overlapping method transfers initialized $C0$ and $C1$ in the beginning~\cite{wang2023overlap}.}. 
As we can see, ideally communication can be completely hidden by GEMM computation.

These methods might work greatly on TPUs, but not on GPUs, due to different programming models. 
First, the performance of these methods heavily rely on the execution order, concurrent execution, and execution timing of independent partitions. 
While the execution order and concurrent execution among GPU kernels can be achieved through streams and events\footnote{
Different GPU programming models might have different 
terminology. 
In this paper, we mainly use CUDA terminology.}, however, the execution timing is not trivially controlled by most GPU programming models.
The time variance might be stable and controllable in per-operation evaluation, 
but it typically becomes unpredictable in the real production environment that involved with many streams and events.  
Second, ReduceScatter overlapping typically requires performing additional computation operations, such as \texttt{add} operations in Figure~\ref{fig:existing_rs}, between GEMM operations, creating data dependence that avoid concurrent execution of multiple GEMM kernels through GPU multiplexing\footnote{GPU multiplexing can be achieved through multiple CUDA streams.}. 
Although the \texttt{add} operations can be further fused with communication~\cite{te}, they still avoid concurrent execution of multiple GEMM kernels. 
Last and most importantly, splitting one single large GEMM kernel to multiple smaller GEMM kernels could highly likely underutilize GPU stream processors (SMs) even with a number of partitions as the number of devices, especially when tensor parallelism scales. 

\subsection{Effective Communication Time and Overlapping Efficiency}
\label{sec:eff_comm}

It is non-trivial to indicate performance of communication overlapping methods.
Overlapping methods typically mix communication with computation, increasing difficulty of directly measuring overlapped time. 
Moreover, splitting a GEMM kernel into multiple small GEMM ones delivers longer computation time, 
but longer computation time might provide a higher chance to overlap communication. 
In the end, it may or may not imply shorter overall time.
Overall time is a fair performance indicator, but different methods might still use different GEMM algorithm and/or implementation, impacting overall time. 

We propose \textbf{Effective Communication Time} in Eq.~\ref{eq:comm_eff} to fairly compare different methods and highlight communication time. 
Effective communication time (\texttt{$ECT$}) is defined as overall time ($OverallTime$) minus with \textbf{best}, \textbf{non-split} GEMM computation time ($GEMM_{non-split}$). 
\begin{equation}
ECT = OverallTime - GEMM_{non-split} 
\label{eq:comm_eff}
\end{equation}
To minimize impacts of GEMM kernels, we use the \textit{fastest} GEMM kernels from the best of our knowledge in all of the evaluation. 
Since we use the same, fastest GEMM kernels across methods, given a shape of problem, making $GEMM_{non-split}$ identical across different methods, effective communication time simply has a shift of overall time, but highlights more in communication. 
Particularly, for a non-overlapping method, effective communication time is equal to regular communication running with the fastest GEMM kernels,
while for an overlapping method, slowdown time from any inefficient factor and non-overlapped communication portion all contributes to its effective communication time.  
It is also worth mentioning that a perfect overlapping method delivers zero effective communication time.

On top of effective communication time, we further define \textbf{Overlap Efficiency} ($E_{overlap}$) in Eq~\ref{eq:overlap_eff}, as 
one minus the ratio between effective communication time of an overlapping method ($ECT_{overlap}$) and effective communication time of an non-overlapping baseline ($ECT_{non-overlap}$).
\begin{equation}
E_{overlap} = 1 - \dfrac{ECT_{overlap}}{ECT_{non-overlap}}
\label{eq:overlap_eff}
\end{equation}
To minimize impacts of the non-overlapping baseline, the widely used, standard GPU communication library, NCCL~\cite{nccl}, is chosen in all of the evaluation, and NCCL is also the fastest non-overlapping communication from the best of our knowledge. 
Particularly, overlap efficiency of the non-overlapping baseline is zero, 
while a perfect overlapping method has a 100\% overlap efficiency. 
An overlapping method with a negative overlap efficiency implies that the method runs slower than the non-overlapping baseline. 

\begin{figure}[t]
\centering
\includegraphics[scale=0.5,bb=0 0 550 450]{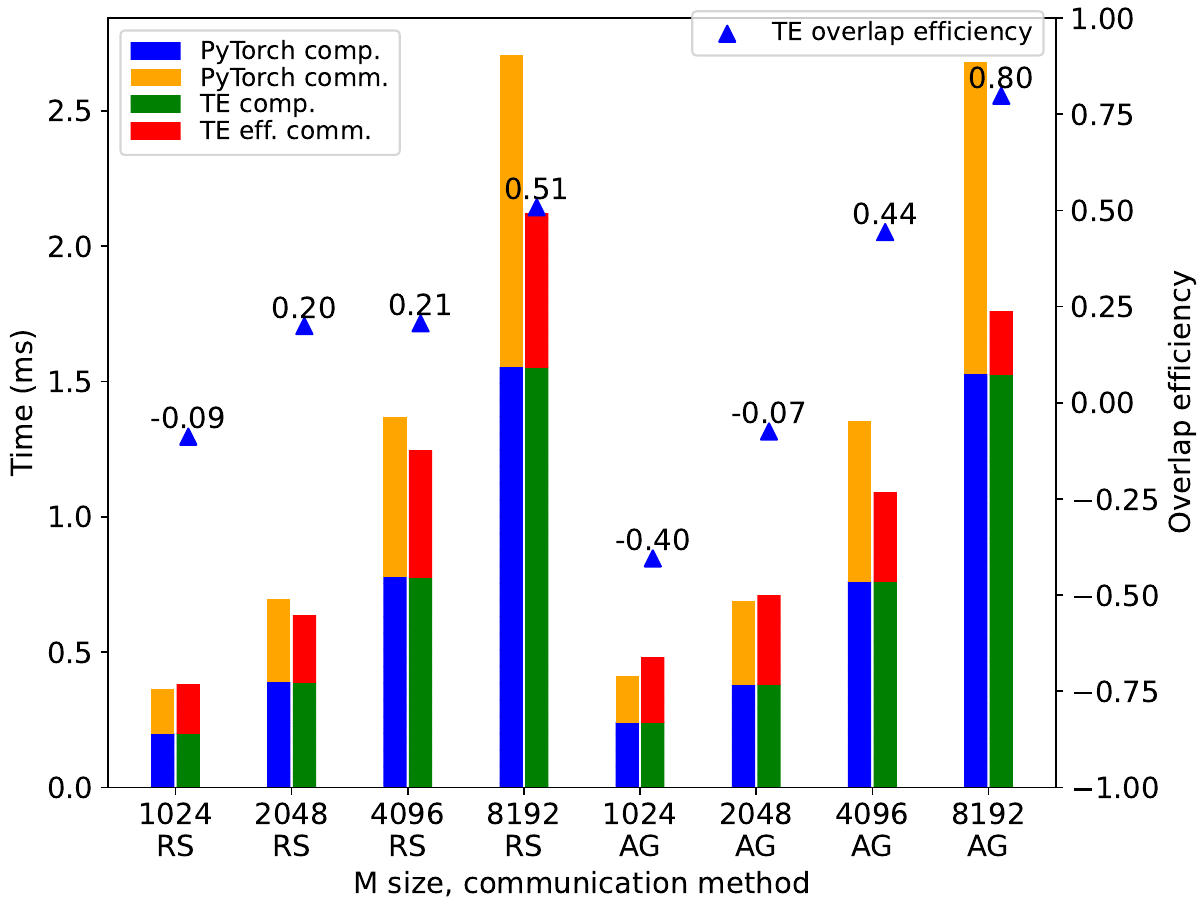}
\caption{Performance between PyTorch (non-overlapping) and TransformerEngine (prior overlapping method) from m = 1024 to 8192, with (n, k) as (49152, 12288) and (12288, 49152) in AllGather and ReduceScatter on an 8-H800 cluster with NVLink interconnections.}
\label{fig:limitation}
\end{figure}

Figure~\ref{fig:limitation} shows computation time and effective communication time of PyTorch and the conventional overlapping technique (implemented with TransformerEngine~\cite{te}), and the corresponding overlapping efficiency, demonstrating a overlapping technique might deliver poor performance or even worse than the original non-overlapping method, due to the above-mentioned limitations. 
When $m$ of the matrix is small, the prior overlapping technique delivers worse performance than the non-overlapping baseline (PyTorch), that supports the third reason mentioned earlier.
The prior overlapping technique performs better in AllGather than ReduceScatter. 
It is mainly because the splited GEMM operations in AllGather might run concurrently through GPU multiplexing, but ReduceScatter cannot, that supports the second reason mentioned earlier. 
\section{Overview of Fused GEMM with Communication}
\label{sec:method}

This paper proposes a more efficient communication overlapping method, \flux, over the conventional methods~\cite{jangda2022breaking, te} on GPUs. 
Different from the existing methods partitioning the computation and communication into the number of devices or twofold of the number,
\flux overdecomposes computation and communication into tiles. 
Here, since the computation operation is GEMM, 
and most high-performance GEMM kernels on GPUs are written with tiling, 
such as thread block tiling or warp tiling, our decomposition can naturally map into existing tiling in the kernels.
\flux fuses dependent communication and/or wait logic into a GEMM kernel, and launches only one fused kernel, compared to the prior methods launching multiple split GEMM kernels.
Considering \flux has much finer-grained than the prior methods, in the rest of the paper, we would refer the prior ones as medium-grained decomposition, and the proposed method as fine-grained decomposition.

\begin{algorithm}[t]
 \KwParameters{Input matrix pointers $A$, $B$}
 \KwParameter{List of output matrix pointers $Cs$}
 \KwParameters{Int scalars $rank\_id$, $N_{TP}$}
 $[m, n] \gets \texttt{TileCoord}(threadblock\_id, rank\_id, N_{TP})$\;
 $ acc \gets 0 $\;
 standard GEMM prologue($A$, $B$, $m$, $n$)\;
 standard GEMM mainloop($A$, $B$, $m$, $n$) updating $acc$\;
 // epilogue start \\
 $C \gets \texttt{GetOutput}(Cs,  N_{TP}, m, n)$\;
 \eIf{fuse reduction}{
   \texttt{Reduce}($C$, $acc$)\;
 }{
   \texttt{Write}($C$, $acc$)\;
 }
 // epilogue end \\
 \caption{A simplified GEMM-ReduceScatter (or -AlltoAll) overlapping kernel}
\label{alg:rs}
\end{algorithm}

\subsection{ReduceScatter Overlapping}
\label{sec:rs}

In \flux, ReduceScatter is implemented as epilogue fusion into a GEMM kernel. 
More specifically, ReduceScatter communication is fused into the epilogue of the GEMM kernel.
Algorithm~\ref{alg:rs} shows pseudocode of the fused GEMM with ReduceScatter (or AlltoAll) for $C = A \times B$, where $A$ and $B$ are the two input matrices, and $Cs$ are a collection of output matrix pointers on all devices involved in tensor parallelism. 
Unlike a standard GEMM kernel having only one single output pointer, the number of output pointers ($Cs$) in the fused GEMM kernel is increased to the number of devices in tensor parallelism ($N_{TP}$), and can be collected through inter-process communication in the initialization phase of the corresponding PyTorch operation.  
The output coordinate ($m$ and $n$) can be computed through a function \texttt{TileCoord} with thread block indices and the local rank index($rank\_id$). 
Selection (\texttt{GetOutput}) of an output pointer in the fused GEMM is based on the output coordinate ($m$ and $n$) and the number of devices in tensor parallelism ($N_{TP}$). 
For example, in the two GEMM operations of Figure~\ref{fig:tp}, the selection is based on the row index.

A ReduceScatter operation can be further decoupled to an \textit{AlltoAll} operation and a \textit{reduction} one.
Here, AlltoAll refers to only communication across devices, while reduction happens locally on individual devices. 
Therefore, fusing AlltoAll (\texttt{Write} branch) into GEMM epilogue is typically enough to overlap communication, the reduction fusion (\texttt{Reduce} branch) only provides marginal performance gain.
Section~\ref{sec:reduce} would discuss the implementation details of reduction. 

This algorithm requires GPUs with peer-to-peer (P2P) supports, which 
modern NVIDIA GPUs within a node already have, regardless with NVLink or PCIe interconnects. 
NVSHMEM~\cite{nvshmem} extends P2P on NVIDIA GPUs across nodes.
The detailed implementations of \texttt{TileCoord}, \texttt{Reduce} and \texttt{Write} would be discussed in Section~\ref{sec:opt}.

\begin{algorithm}[t]
 \KwParameters{Input matrix pointers $A_{agg}$, $B$}
 \KwParameter{Output matrix pointer $C$}
 \KwParameter{List of scalar $signal\_list$}
\KwParameters{Int scalars $rank\_id$, $N_{TP}$}
 $[m, n] \gets \texttt{TileCoord}(threadblock\_id, rank\_id, N_{TP})$\;
 $signal \gets \texttt{GetSignal}( signal\_list,  N_{TP}, m, n)$\;
  \texttt{WaitSignal}($signal$)\;
 standard GEMM($A$, $B$, $C$, $m$, $n$)\;
 \caption{A simplified AllGather-GEMM overlapping kernel }
\label{alg:ag}
\end{algorithm}

\begin{algorithm}[t]
 \KwParameter{List of input matrix pointers $A\_list$}
 \KwParameter{List of output matrix pointer $A_{agg}\_list$}
 \KwParameter{List of scalar $signal\_list$}
 \KwParameter{Int scalar $rank\_id$, $N_{TP}$}
 \KwParameter{List of tile info $tiles_{comm}$}
 \For {$tile$ from  $tiles_{comm}$} {
    \eIf{pull} 
    {
        // pull-based \\
        $A_{remote} \gets \texttt{GetRemotePtr}(A\_list, tile)$\;
        $A_{local} \gets \texttt{GetLocalPtr}(A_{agg}\_list, tile)$\;
        $\texttt{DataTransfer}(A_{remote}, A_{local}, tile.size)$\;
    } {
        // push-based \\
        $A_{remote} \gets \texttt{GetRemotePtr}(A_{agg}\_list, tile)$\;
        $A_{local} \gets \texttt{GetLocalPtr}(A\_list, tile)$\;
        $\texttt{DataTransfer}(A_{local}, A_{remote}, tile.size)$\;
    }
    $signal \gets \texttt{GetSignalHost}(signal\_list, tile)$\;
    \texttt{SetSignal}($signal$)\;
 }
 \caption{A host function for AllGather-GEMM overlapping}
\label{alg:ag_host}
\end{algorithm}

\subsection{AllGather Overlapping}
\label{sec:ag}

\begin{figure}[t]
\centering
\includegraphics[scale=0.7,bb=0 0 450 450]{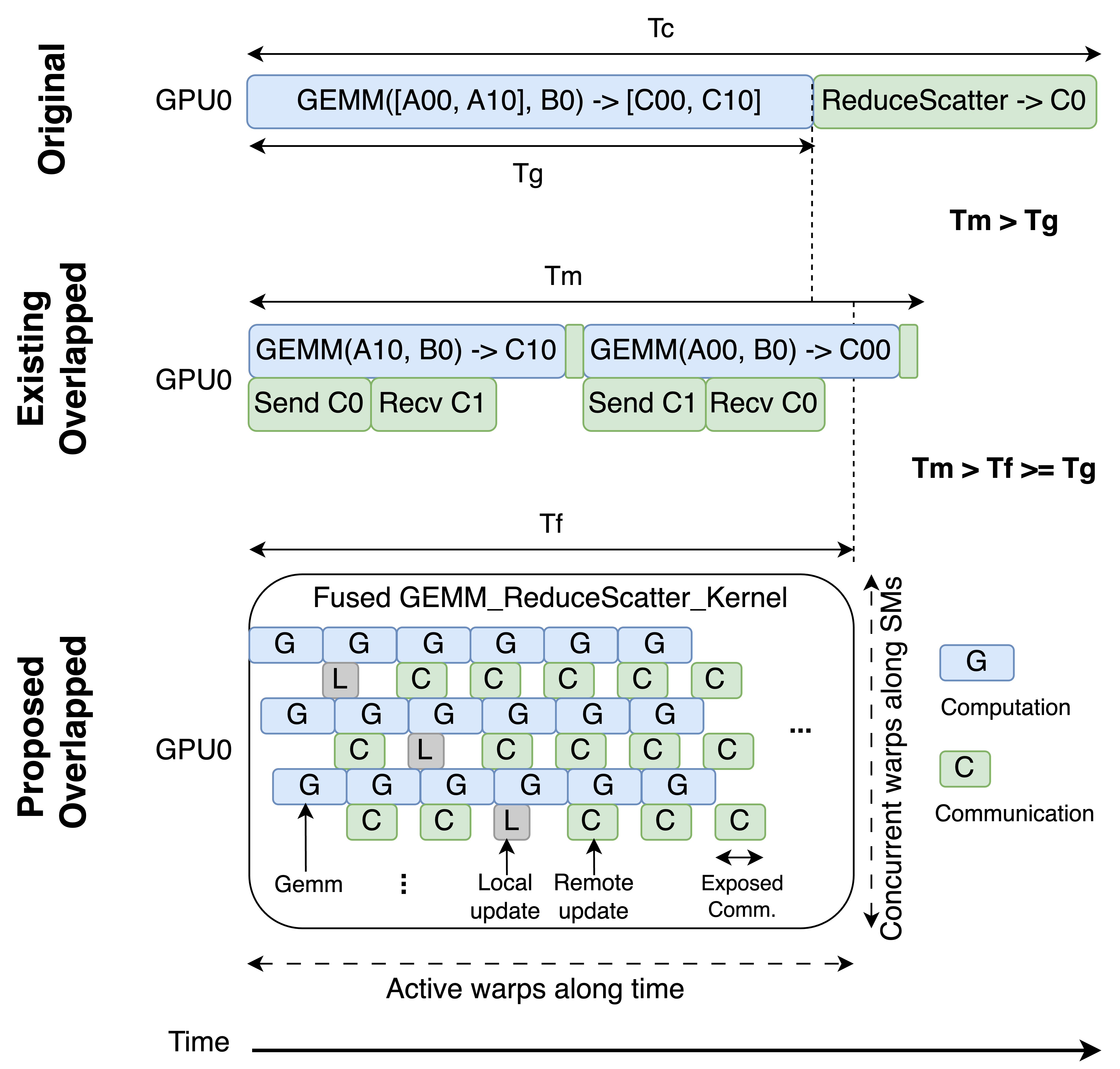}
\caption{An illustration of differences among the non-overlapping and different overlapping methods in a GEMM-ReduceScatter pattern with 2-way tensor parallelism.}
\label{fig:compare_rs}
\end{figure}

Different from ReduceScatter, AllGather is implemented as prologue fusion into a GEMM kernel. 
More specifically, AllGather signal checking is fused into the prologue of the GEMM kernel.
Algorithm~\ref{alg:ag} shows pseudocode of the fused GEMM with AllGather for $C = A_{agg} \times B$,  
where $A_{agg}$ is the aggregated matrix buffer for input A matrices through AllGather, $B$ is the other input matrix, and $C$ is the output matrix, 
and Algorithm~\ref{alg:ag_host} shows the corresponding communication happening on the host side.

On the kernel side, GEMM tile computation is blocked by the function \texttt{WaitSignal} until the value contained in the $signal$ is set to true.
Here, the $signal$ is chosen by \texttt{GetSignal} from a collection of signals ($signal\_list$) based on the output coordinate ($m$ and $n$), and the number of devices in tensor parallelism ($N_{TP}$). 
For example, in the MLP of Figure~\ref{fig:tp}, the selection is based on the row index. 
The $signal$ for each communication is only set to true on the host side when the corresponding portion (communication tile) of the input tensor becomes ready, meaning the portion is received on the device running the fused kernel. 

The host side (either pull- or push-based) performs tiled communication operations (\texttt{DataTransfer}) and set the corresponding signals (\texttt{SetSignal}) to true asynchronously.
Particularly, the pull-based method transfers tiles by pulling tiles from remote devices through \texttt{GetRemotePtr} function and \texttt{GetLocalPtr} function choosing the right pointers from a list of the sharded A matrices, $A\_list$, and a list of aggregated matrix buffers, $A_{agg}\_list$, and then setting \textit{local} signals.
The $signal$ is chosen by \texttt{GetSignalHost} from a collection of signals ($signal\_list$) based on the communication $tile$ index. 
On the other hand, the push-based one transfers tiles by pushing tiles to remote devices and then setting \textit{remote} signals. 
Note $signal\_list$ in the pull-based version contains only local signals, while $signal\_list$ in the push-based version contains signals in remote devices.
Selection between these two variants is considered as a tuning knob, and is discussed in Section~\ref{sec:ag_details}. 

It is worth mentioning that in AllGather our method fuses only the wait logic of communication into the GEMM kernel, instead of entire communication operations.
Therefore, AllGather does not necessarily require P2P.
Meanwhile, in AllGather, the tiling strategy of communication ($tiles_{comm}$) is decoupled from the tiling strategy of GEMM computation. 
This design provides a flexible way to choose a trade off between overlapping opportunity and communication efficiency without compromising the GEMM efficiency. 
Section~\ref{sec:opt} would discuss all optimizations, and implementation details in the functions \texttt{TileCoord}, \texttt{WaitSignal}, \texttt{SetSignal}, and \texttt{DataTransfer}.

\begin{figure}[t]
\centering
\includegraphics[scale=0.6,bb=0 0 450 600]{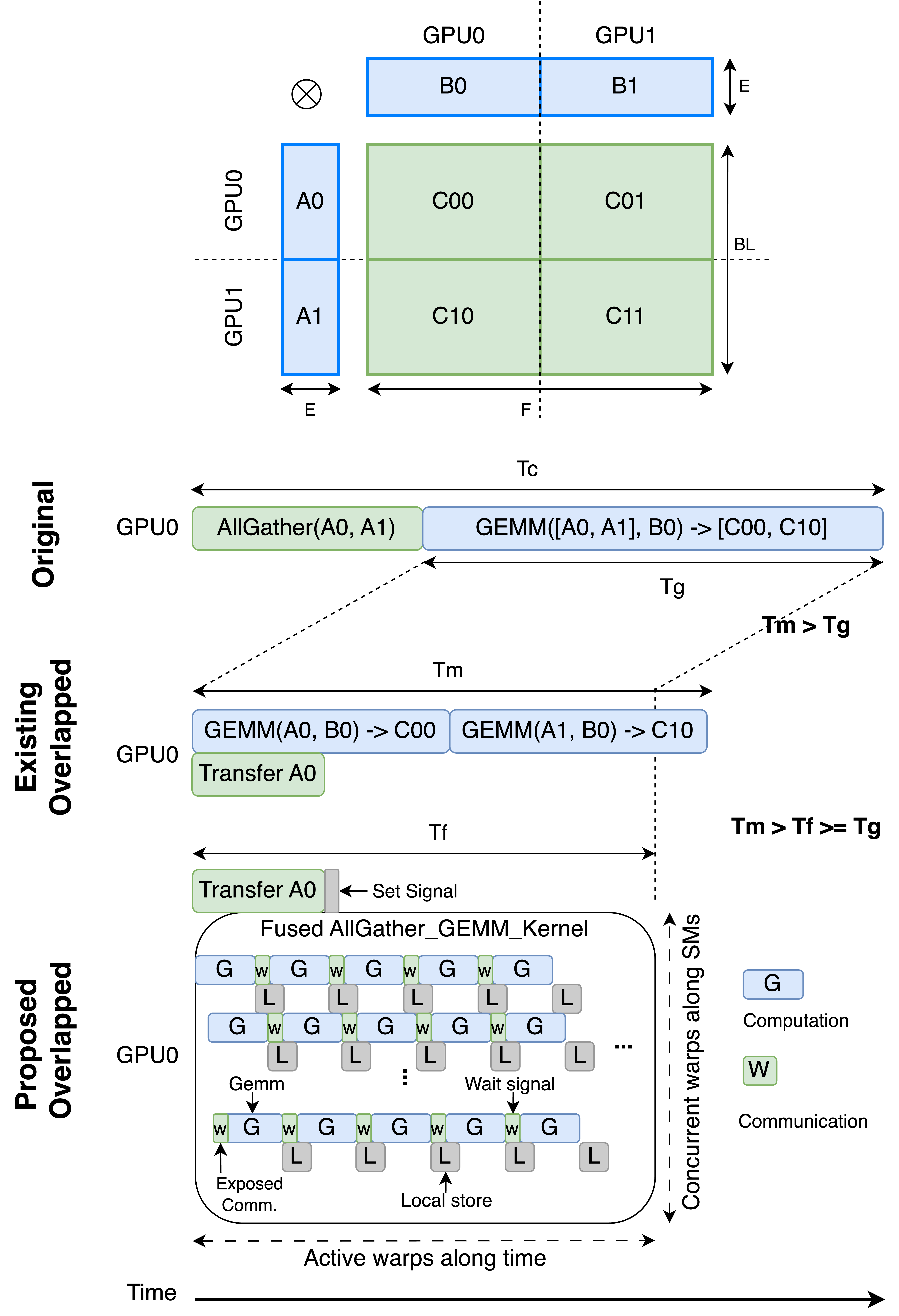}
\caption{An illustration of differences among the non-overlapping and different overlapping methods in an AllGather-GEMM pattern with 2-way tensor parallelism.
}
\label{fig:compare_ag}
\end{figure}

\subsection{Comparison among Decomposition Strategies}
\label{sec:compare}
Figure~\ref{fig:compare_rs} illustrates the major differences among the overlapping techniques in ReduceScatter.  
Although the existing overlapping ($T_m$) can potentially perform faster than the original coarse-grained method ($T_c$), the existing method ($T_m$) is typically still slower than the original GEMM time ($T_g$). 
One major reason is that GPU GEMM efficiency decreases by splitting a GEMM kernel into a sequence of multiple smaller GEMM kernels.
GEMM typically requires reasonably large matrices to fully utilize GPU compute power. 
The sequence of smaller GEMM operations with data dependence further blocks those GEMM kernels from concurrently running through GPU multiplexing, and consequently, the more way of tensor parallelism, the worse GEMM efficiency on GPUs.
Compared to the existing method, our proposed technique does not have the above limitation. 
Our new overlapping technique ($T_f$) can perform as fast as the original GEMM operation ($T_g$) with a very small overhead.
Its fine-grained decomposition strategy perfectly fits the nature of the modern GPU design, latency hiding among context-switching warps and hundreds of concurrent active warps among SMs, illustrated in the bottom zoom-in view. 
In the end, our method only exposes a small portion of communication in the tail of execution without compromising GEMM computation efficiency.

Figure~\ref{fig:compare_ag} illustrates the major differences among the overlapping techniques in AllGather.
Similarly, the existing overlapping ($T_m$) could be faster than the original coarse-grained method ($T_c$), but is still slower than the original GEMM time ($T_g$), due to 
lower GPU GEMM efficiency, and our new overlapping technique ($T_f$) can deliver a similar performance as the original GEMM operation ($T_g$).
The long latency instruction in AllGather is from waiting signals, happening in the beginning of each warp since \texttt{WaitSignal} is fused in the prologue. 
Its latency varies based on arrival time of corresponding data transfers.
For the tile with data already arrived, the latency is close to zero. 
For the tile with data not ready, context switching among warps can hide the latency. 
It is worth mentioning that signals for local tiles are always preset to true, so there is always some warps not needing to wait signals. 
In the end, our method only exposes a small portion of communication in the head of execution without compromising GEMM computation efficiency.
Section~\ref{sec:opt} further discusses optimization reducing the waiting latency.

\begin{figure}[t]
\centering
\includegraphics[scale=0.6,bb=0 0 450 300]{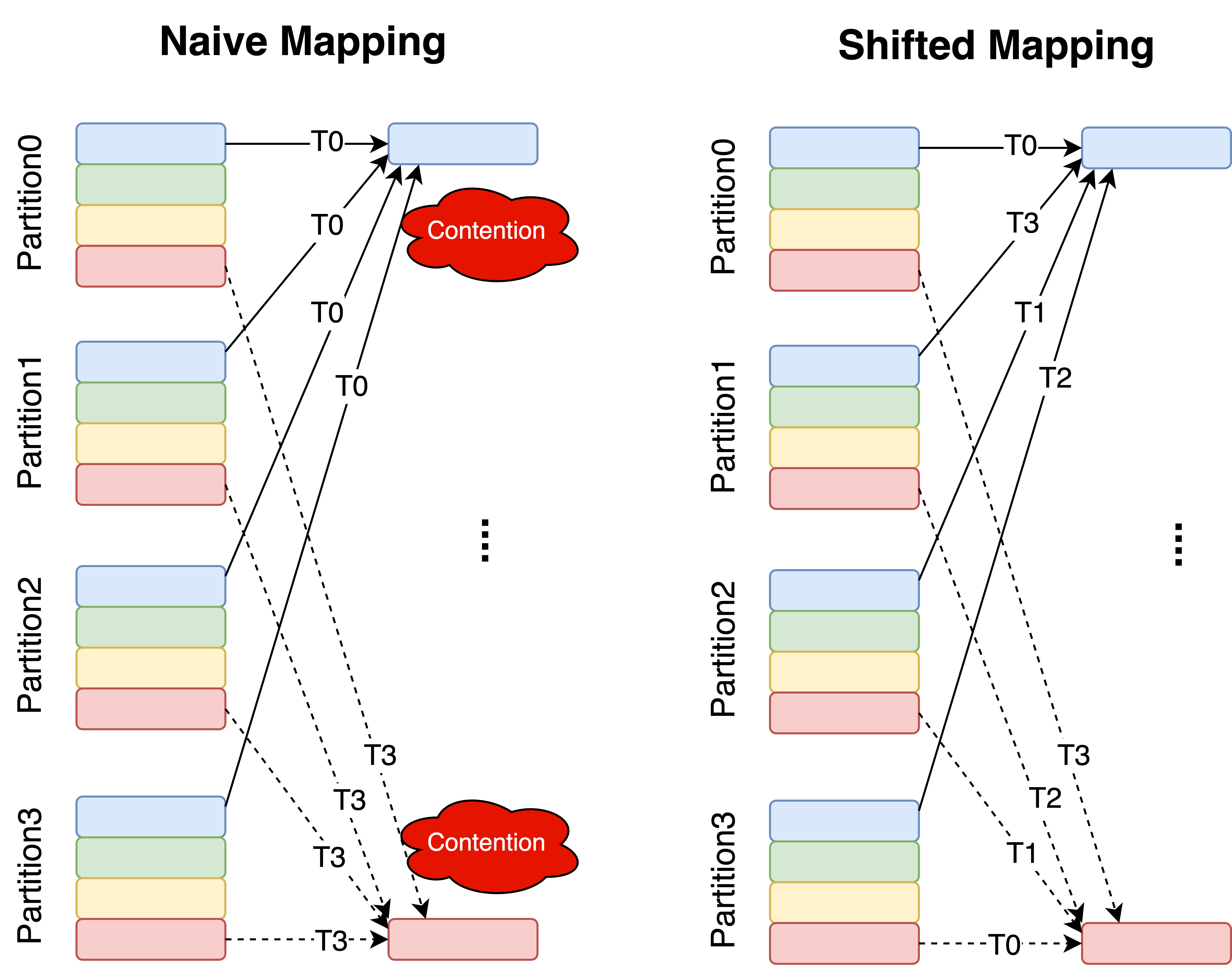}
\caption{An illustration of memory contention happening at time-step $T_i$ in the naive tile coordinate mapping, and the proposed solution in GEMM-ReduceScatter overlapping with 4-way tensor parallelism.}
\label{fig:tile_schedule_rs}
\end{figure}

\section{Optimizations and Implementation Details}
\label{sec:opt}

As mentioned in Section~\ref{sec:compare}, our algorithms fit the nature of the modern GPU design. 
Therefore, direct implementations of Algorithm~\ref{alg:rs},~\ref{alg:ag}, and~\ref{alg:ag_host} can already outperform the prior methods by delivering better communication overlapping and GEMM efficiency. 
This section covers advanced optimizations that push the performance to the limit, and the implementation details.

\subsection{Tile Coordinate Swizzling}
\label{sec:tile_schedule}

An efficient GPU kernel relies on tiling to exploit parallelism and locality. 
Therefore, the kernel has a tile mapping logic, such as \texttt{TileCoord} in Algorithm~\ref{alg:rs} and~\ref{alg:ag}, from a thread block index to a tile coordinate. 
Inspired from a well-tuned GEMM typically swizzling the mapping logic for maximizing memory efficiency~\cite{cutlass},
we explore tile coordinate swizzling to further improve the efficiency of our fused kernels. 

In the fused GEMM-ReduceScatter, tile coordinate is shifted with the device rank index to avoid write request conflicts from the kernels running on different devices, minimizing possible contention in the memory controller on each individual device. 
Figure~\ref{fig:tile_schedule_rs} illustrates the possible memory contention in a naive mapping, and how a shifted mapping avoids the possible memory contention. 

\begin{figure}[t]
\centering
\includegraphics[scale=0.5,bb=0 0 600 450]{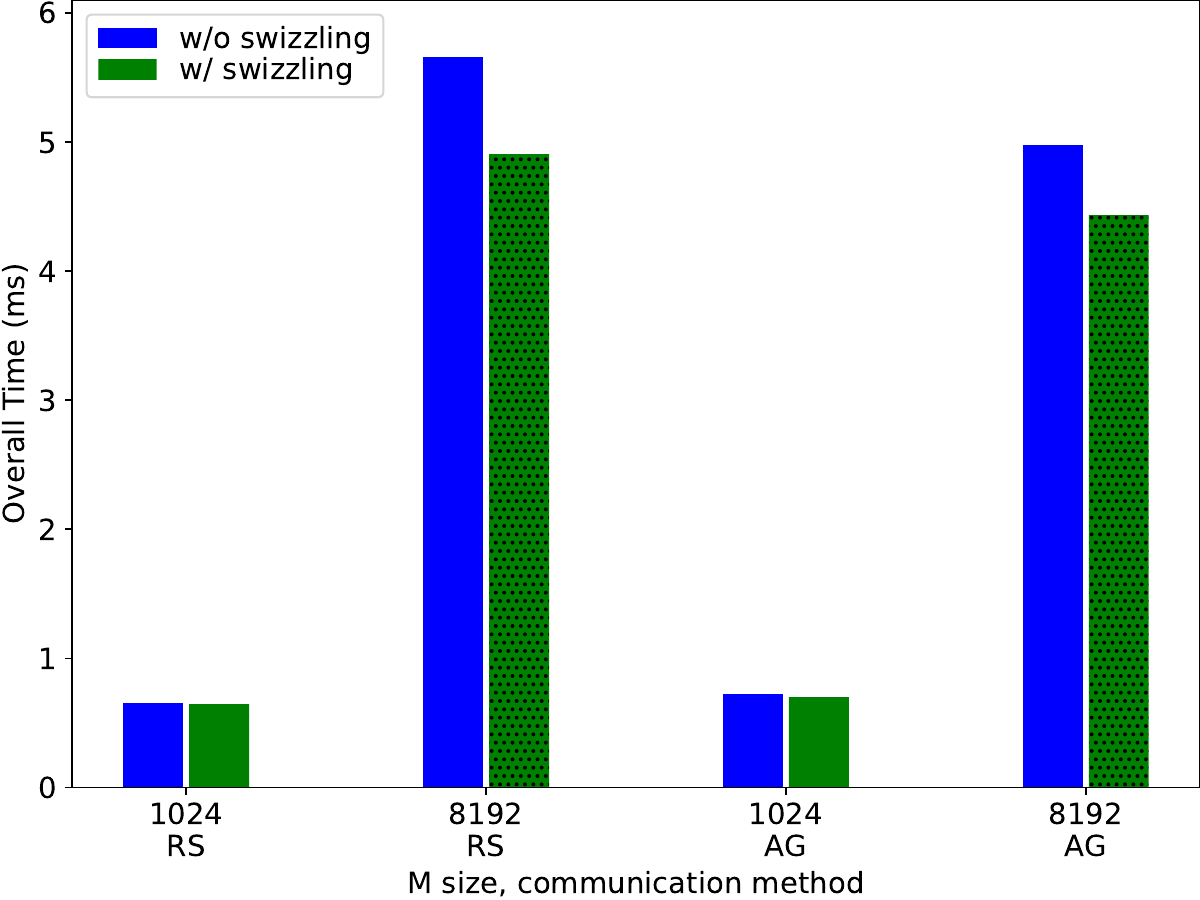}
\caption{Performance with or without applying tile coordinate swizzling for small (1024) and large (8192) m with (n,k) as (49152, 12288) and (12288, 49152) in AllGather and ReduceScatter, respectively, on an 8-A100 NVLink cluster. }
\label{fig:tile_schedule_result}
\end{figure}

A similar strategy is applied in the fused AllGather-GEMM as well to minimize thread blocks waiting, in the end minimizing overall delay. 
The fused AllGather-GEMM requires tile coordinate swizzling (\texttt{TileCoord}) to align with the order of the signal arrival order, which is determined by the communication order on the host side (controlled by $tiles_{comm}$ in Algorithm~\ref{alg:ag_host}). 
In the implementation, these two orders are chosen together based on the network topology to minimize the overall delay, and the more detailed implementation is discussed in Section~\ref{sec:ag_details}.

Figure~\ref{fig:tile_schedule_result} shows performance impacts before and after applying the tile coordinate swizzling technique on an 8-A100 cluster with NVLink interconnects. 
The adjusted mapping with tile coordinate swizzling always outperforms the naive mapping. 
We can also observe that the performance impact increases when the matrix size increases. 
It is mainly because the memory contention of the naive mapping in GEMM-ReduceScatter and the waiting time of AllGather-GEMM also increase when the matrix size increase. 

It is worth mentioning that existing method~\cite{wang2023overlap} also applied a similar swizzling idea by changing execution order of split GEMM operations. 
Since our method does not split a GEMM operation, their idea cannot be directly applied in our algorithms.

\begin{figure}[t]
\centering
\includegraphics[scale=0.5,bb=0 0 600 450]{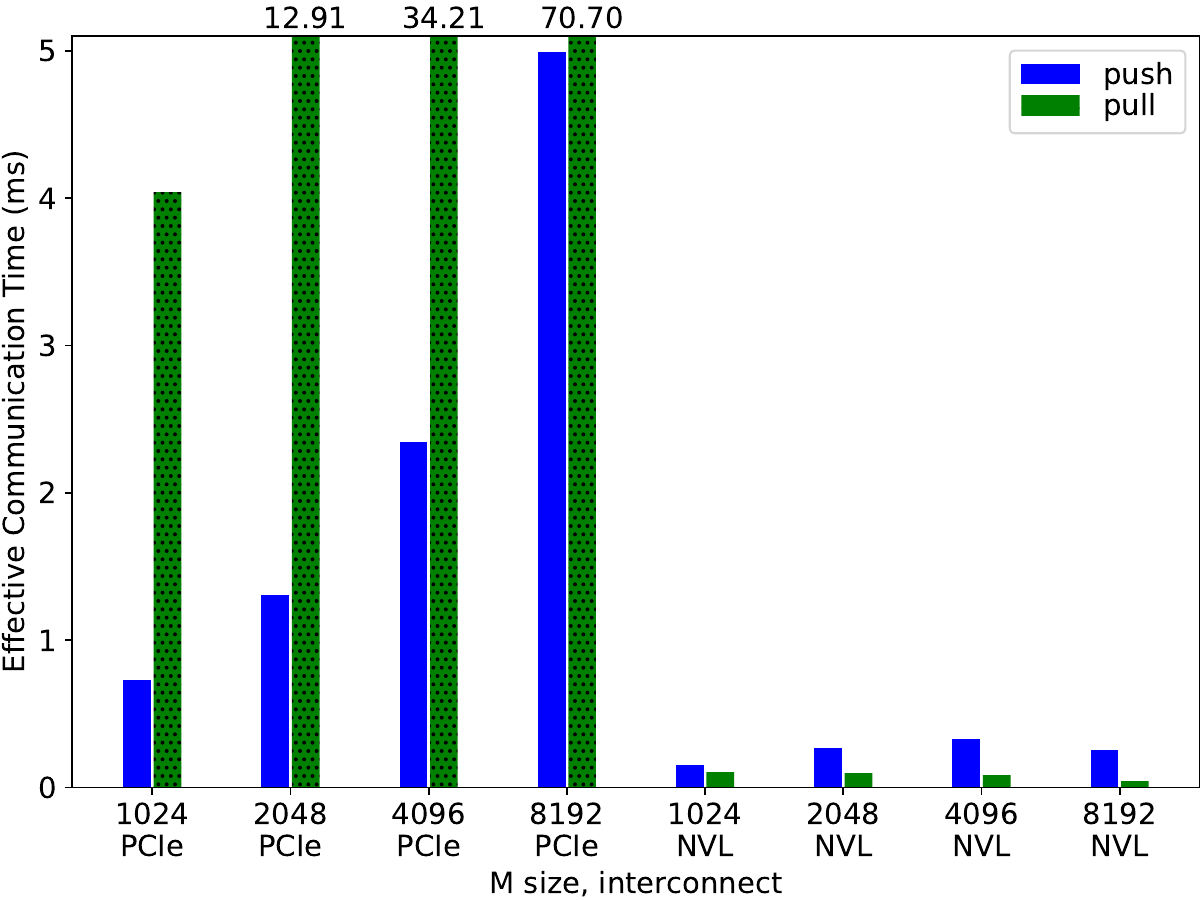}
\caption{Performance comparison between pull- and push-based data transfers with different m, and (n,k) as (49152, 12288) in AllGather on an 8-A100 cluster with PCIe or NVLink interconnects.}
\label{fig:pull}
\end{figure}

\subsection{Implementation Details of ReduceScatter}
\label{sec:reduce}

\textbf{Write.}
Writing data on a local GPU or remote intra-node P2P GPUs is implemented through 
1) storing data from registers to global memory using all variants of \texttt{st} instructions (including vector versions), 
2) storing data from scratchpad to global memory using all variants of \texttt{cp.async.bulk} instructions or 
all variants of Tensor Memory Accelerator (TMA) instructions \texttt{cp.async.bulk.tensor} on Hopper GPUs. 
On the other hand, for writing data on remote inter-node GPUs, NVSHMEM is applied and those writes are implemented through all variants of \texttt{put} APIs. 
All methods are implemented using CUTLASS EVT~\cite{evt} with templates, and template parameters are chosen during auto-tuning.

\textbf{Reduce.}
As mentioned in Section~\ref{sec:rs}, reduction can be potentially fused into the GEMM kernel as well. 
In this case, 1) \texttt{red} or atomic instructions can be used to directly implement reduction on device memory without changing code structure or introducing too much overhead if GPUs enable P2P memory access. These kinds of instructions are useful, but might not support all data types or all kinds of GPUs\footnote{BF16 atomic operations are not supported on A100 and H800.}. Therefore, we only apply these instructions for selected data types with capable GPUs.
On Hopper GPUs, 2) warp or thread block \textit{specialization}\footnote{Warp or thread block specialization is a CUDA programming method on Hopper GPUs allowing a warp or thread block in a kernel to perform a specific task, such as load/store/wgmma and synchronizing warps or thread blocks performing different tasks within a single kernel in a producer-consumer fashion.}~\cite{cutlass} is applied to implement reduction by each GPU writing partial results to its local memory and a specialized warp or thread block pulling ready remote data to perform a local reduction on the destination GPU. 
These kinds of warp or thread block specialized reduction methods specifically perform well with warp or thread block specialized GEMM kernels on Hopper. 
For remote inter-node GPUs, we fuse only AlltoAll in the kernel, and perform discrete reduction. 
All methods are also implemented using CUTLASS EVT with templates, and template parameters are chosen during auto-tuning.

\subsection{Implementation Details of AllGather}
\label{sec:ag_details}

\textbf{DataTransfer.}
Although the proposed AllGather algorithm does not necessarily require P2P, 
we still separate implementations with and without P2P. 
For GPUs with P2P memory access, either pull-based or push-based transfers can be implemented with \texttt{cudaMemcpy} APIs.
The only differences are the pointers. Pull-based uses a local destination pointer and a remote source pointer, while push-based uses the opposite way. 
Figure~\ref{fig:pull} shows performance difference between two transfer methods on 8-A100 PCIe and 8-A100 NVLink clusters.
As the results shown, different interconnects might have different preference. 
Therefore, autotuning is applied to select proper transfer methods.
On the other hand, for GPUs without P2P access, NCCL~\cite{nccl} send/recv are used. Since NCCL send/recv are paired, there is not difference between pull or push.
All methods are implemented in C++ with templates, and template parameters are chosen during auto-tuning.

\textbf{Signals.}
We use a regular 32-bit GPU memory to implement a signal. All signals are allocated contiguously for easy preset and reset, and they are preset in the corresponding PyTorch operator's initialization 
or reset after the corresponding GEMM computation with a stream and an event avoiding data race.  
On the host side, a signal is set through a \texttt{cuStreamWriteValue} API with a stream, while on the kernel, \texttt{WaitSignal} is implemented through \textit{spinning}.

\begin{figure}[t]
\centering
\includegraphics[scale=0.5,bb=0 0 600 450]{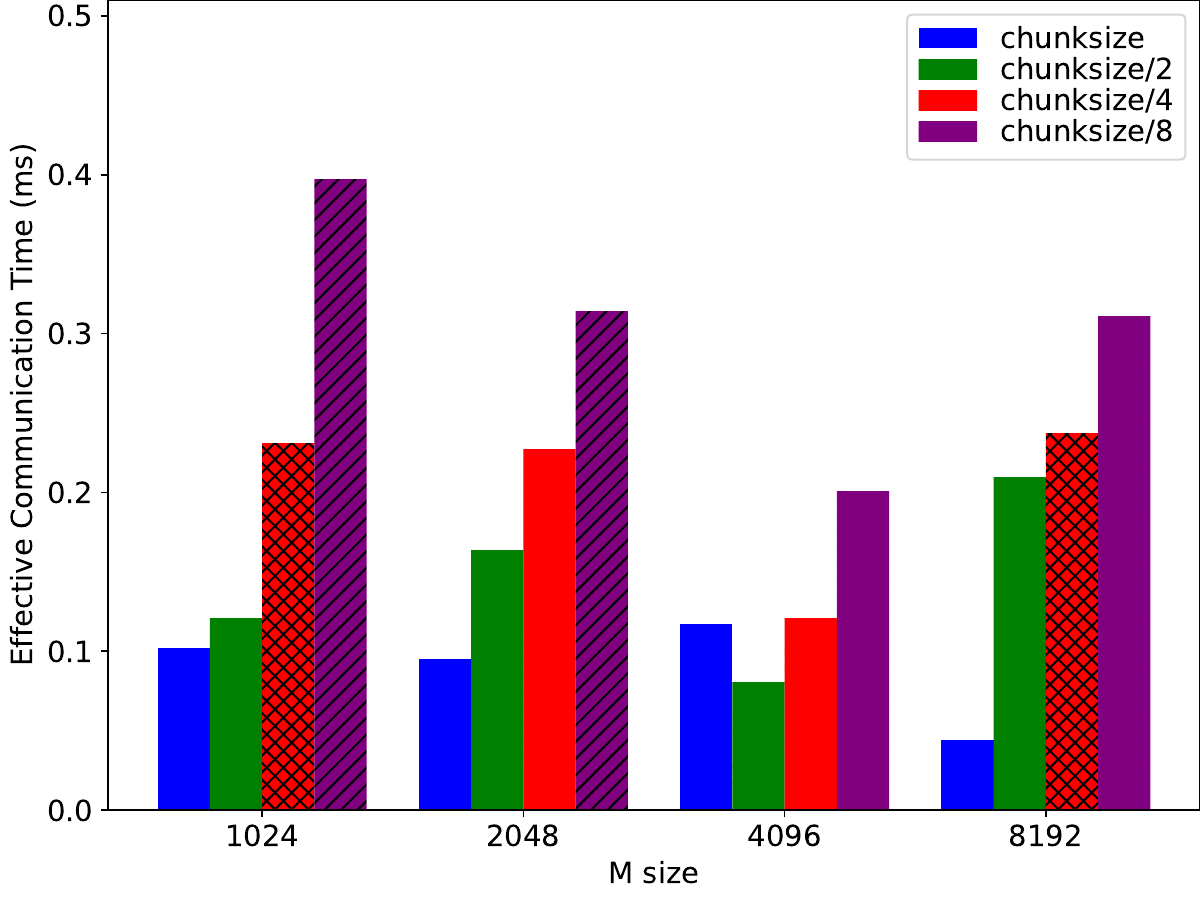}
\caption{Performance results among different communication tile sizes with different m, and (n,k) as (49152, 12288) in AllGather on an 8-A100 NVLink cluster.}
\label{fig:split}
\end{figure}

\textbf{Communication tile size.}
In AllGather, communication tiling is decoupled from GEMM computation tiling for avoiding interfering GEMM tiling, considering GEMM performance is sensitive to GEMM tiling.  
Tuning communication tiling independently allows us to find a best trade off between overlapping opportunity and communication efficiency, minimizing effective communication time. 
During tuning, we start from the tiling size of medium-grained partitioning (denoted as chunksize in Figure~\ref{fig:split}), meaning the tiling size equal m divided by the number of tensor parallelism, and then keep dividing by two until it equal to the GEMM tile size.
Figure~\ref{fig:split} shows the communication tile size does impact the overall performance. However, since there is no clear trend that one size always outperforming the other, 
autotuning is applied to select a best tiling factor.

\textbf{Communication order among tiles.}
As discussed in Section~\ref{sec:tile_schedule}, the communication order on the host side is aligned with tile coordinate swizzling, and chosen based on the network topology to minimize the overall delay.
Intra-node NVLink interconnects apply direct communication with a ring order starting after the local rank. For example, given a local rank index $5$ of 8-way tensor parallelism, the communication order of this rank is 6, 7, 0, 1, 2, 3, 4. 
Intra-node PCIe interconnects use ring-based communication to efficiently utilize PCIe bandwidth for single-node tensor parallelism. 
In multi-node tensor parallelism, for example, 16-way tensor parallelism, inter-node communication \textit{potentially} can overlap with intra-node communication as well.
Therefore, in the intra-node NVLink interconnects, inter-node direct communication is issued with local intra-node communication, and then after each communication tile received from inter-node communication, corresponding new intra-node communication would be issued.
In the intra-node PCIe interconnects, communication is much tricky, since some parts of PCIe interconnects are shared between inter-node and intra-node communication. 
In the evaluated PCIe cluster (the A100 PCIe cluster in Section~\ref{sec:eval}), 4 GPUs and 1 NIC connect to one CPU core, and there are 2 CPU cores per node.  
In this cluster, inter-numa (still intra-node) communication and inter-node communication should not be scheduled at the same time for avoiding possible traffic. 
Therefore, inter-numa communication is issued first, and then intra-numa and inter-node communication is issued together.

\subsection{GEMM Implementation and Auto-Tuning}
\flux is generally applicable for almost all kinds of GEMM kernels. 
Considering GEMM performance is critical for overall performance, workload-balanced GEMM~\cite{osama2023stream} is typically preferred on Ampere GPUs, and warp or thread block specialized GEMM~\cite{cutlass} is preferred on Hopper GPUs. 
Also, since regular tiling of GEMM in \flux is not bond to the number of tensor parallelism, tiling sizes can be adjusted without impacting correctness. 
\flux is implemented using CUTLASS~\cite{cutlass} to fully control GEMM tiling and corresponding prologue or epilogue fusion.  
Similarly to traditional GEMM libraries tuning and selecting GEMM kernels based on matrix shapes, data types, and GPU architecture,
all prologues, epilogues, GEMM algorithms, and all tuning knobs, are written in templates, allowing us to autotune kernels by selecting proper template parameters.

\section{Evaluation}
\label{sec:eval}

\flux is implemented with CUTLASS 3.4.1~\cite{cutlass} and NVSHMEM 2.10.1~\cite{nvshmem}, with compiled with NVCC 11.8 for NVIDIA A100 GPUs and NVCC 12.2 for H800 GPUs.
The results are evaluated with bfloat16 on three different clusters, 
1) an A100 PCIe (80GB) cluster (denoted as A100 PCIe) with 8 GPUs per node, PCIe intra-node interconnects, and 2 100Gbs inter-node interconnects (4 GPUs and 1 NIC per CPU core), 
2) an A100 SXM4 (80GB) cluster (denoted as A100 NVLink) with 8 GPUs per node, NVLink intra-node interconnects, and 4 200Gbs inter-node interconnects (2 GPUs sharing 1 200Gbs inter-node interconnect),
and 3) an H800 SXM5 cluster (denoted as H800 NVLink) with 8 GPUs per node, NVLink intra-node interconnects, and 8 400Gbs inter-node interconnects (each GPU having its own dedicated 400Gbs inter-node interconnect to its corresponding GPUs on other nodes). 

For the existing medium-grained overlapping method, we use TransformerEngine 1.4.0~\cite{te} with UserBuffer, since the rest~\cite{jangda2022breaking, wang2023overlap} are not available\footnote{\cite{jangda2022breaking} does not support well new GPUs, like A100 or H800, while~\cite{wang2023overlap} only supports TPUs.} for A100 and H800 GPUs. 
TransformerEngine has multiple configurations for communication overlapping, and the reported numbers are from the best of all configurations.  
In operation-level evaluation, we evaluate GEMM with ReduceScatter, and AllGather patterns, and report both computation time and effective communication time, as well as overlap efficiency (defined in Section~\ref{sec:eff_comm}), while in model-level evaluation, we evaluate the entire model and only report overall time.

\begin{figure}[t]
\centering
\includegraphics[scale=0.4,bb=0 0 600 550]{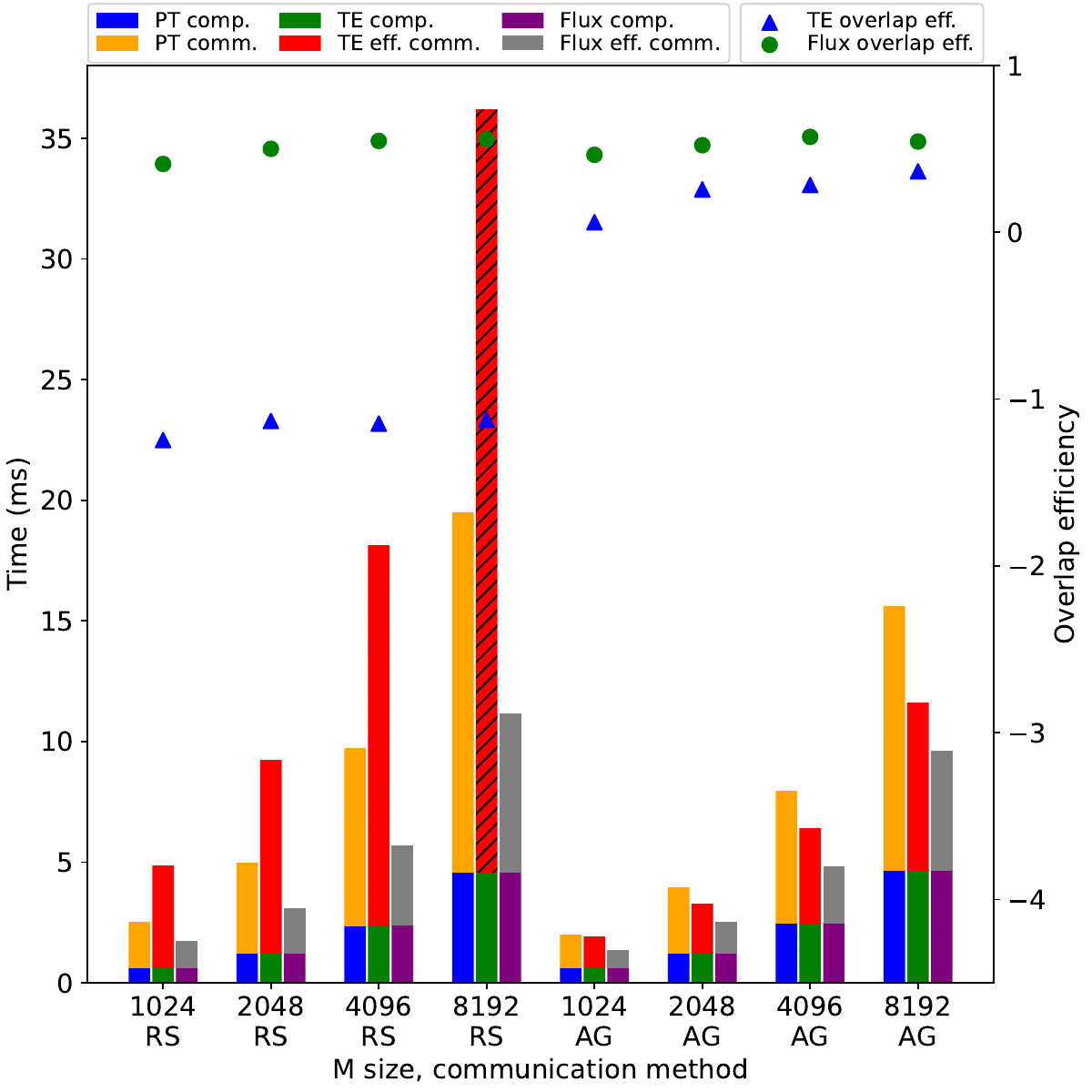}
\caption{Performance results on an 8-A100 PCIe cluster}
\label{fig:pcie_8tp}
\end{figure}

\begin{figure}[t]
\centering
\includegraphics[scale=0.4,bb=0 0 600 550]{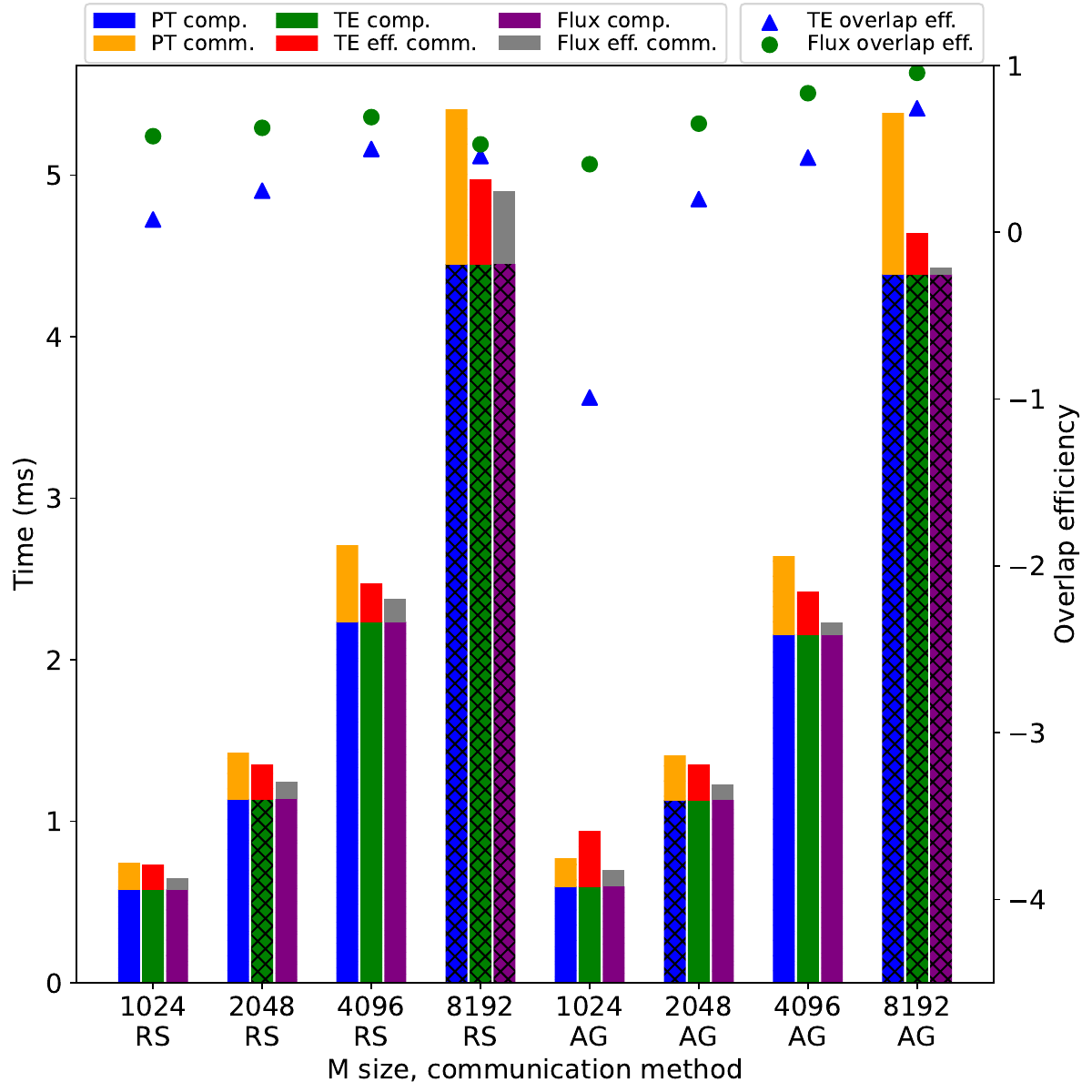}
\caption{Performance results on an 8-A100 NVLink cluster}
\label{fig:a100_8tp}
\end{figure}

\begin{figure}[t]
\centering
\includegraphics[scale=0.4,bb=0 0 600 550]{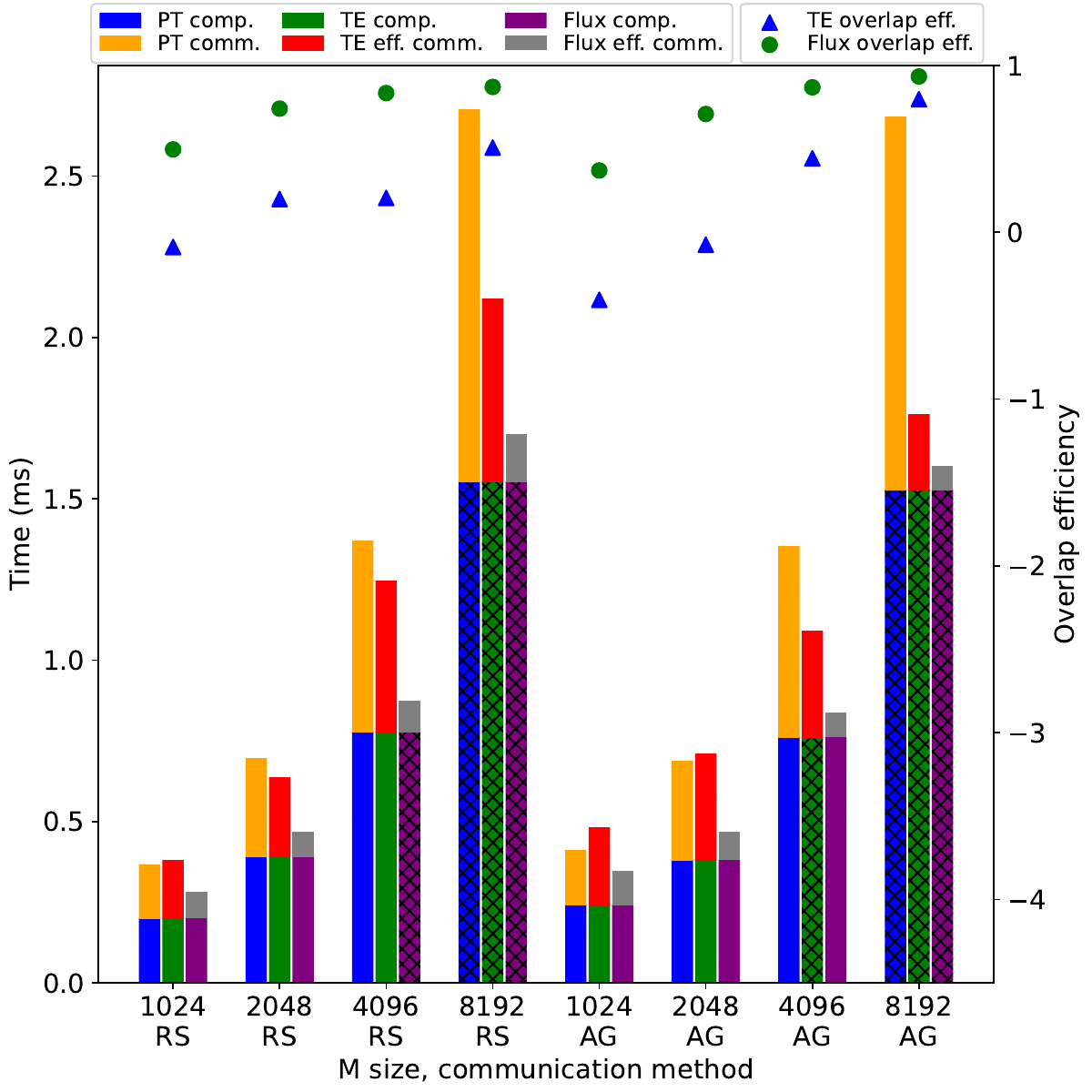}
\caption{Performance results on an 8-H800 NVLink cluster}
\label{fig:h800_8tp}
\end{figure}

\begin{figure}[t]
\centering
\includegraphics[scale=0.4,bb=0 0 700 680]{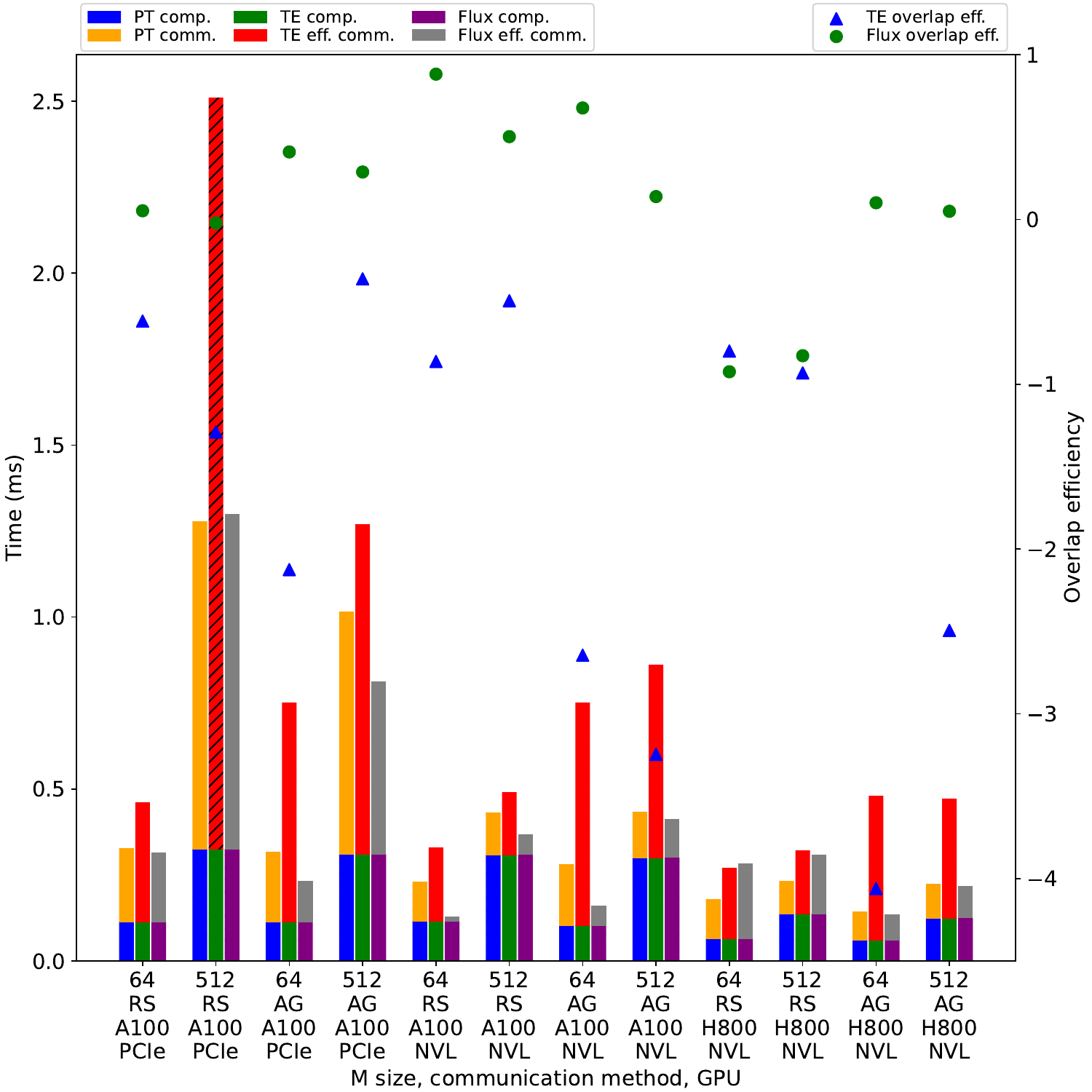}
\caption{Performance results for small m sizes}
\label{fig:small_8tp}
\end{figure}

\begin{figure}[t]
\centering
\includegraphics[scale=0.4,bb=0 0 600 400]{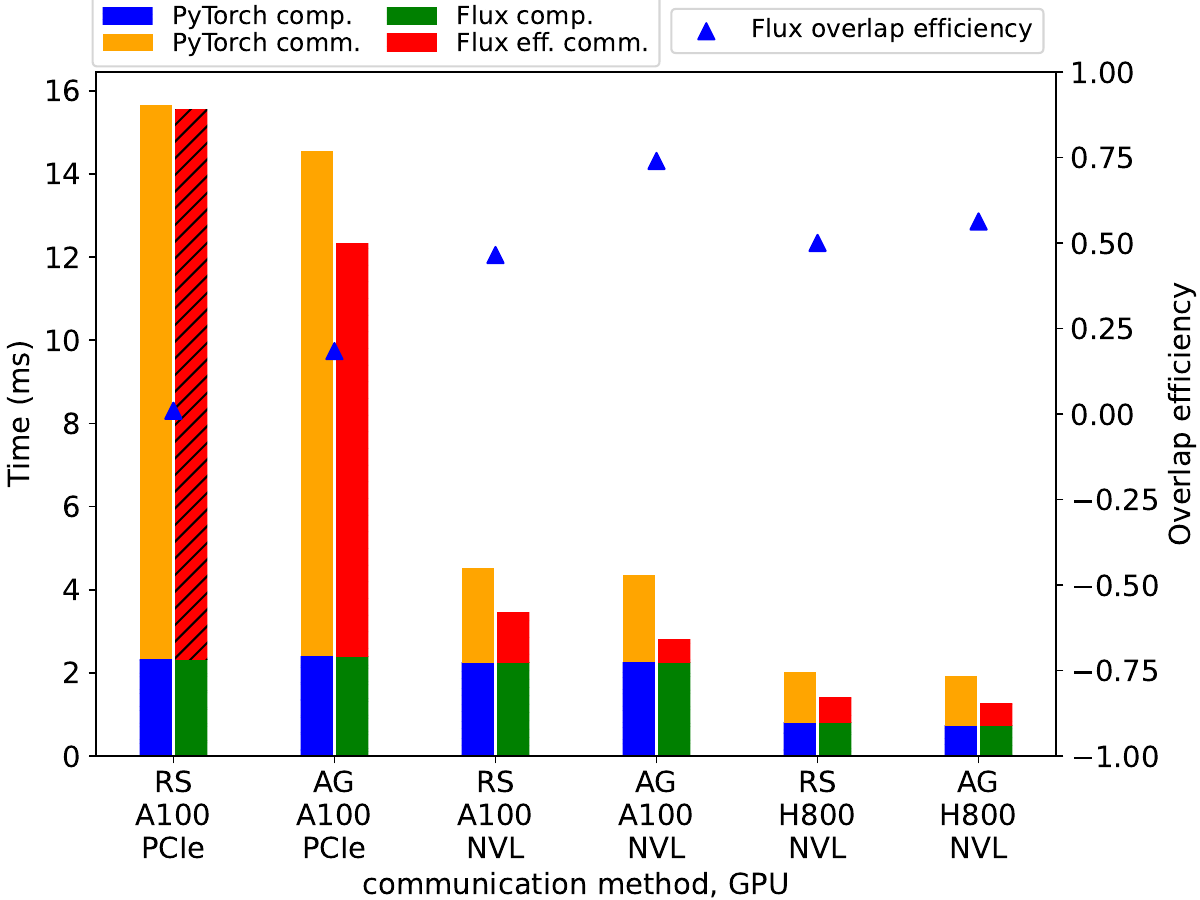}
\caption{Performance results for 16-way tensor parallelism}
\label{fig:tp16}
\end{figure}

\subsection{Operation-level Performance Evaluation}

The GEMM dimensions for evaluation are selected from GPT-3 175B~\cite{brown2020language}.
Therefore, (n, k) is determined as (49152, 12288) and (12288, 49152) in AllGather and ReduceScatter, respectively. 
Note we use (n, k) is the original shape before applying tensor parallelism.
We evaluated GEMM for m from 1024 to 8192, simulating different workloads on training and prefill phases, and much smaller m as 64 and 512 for workloads on decoding phases. 

Figure~\ref{fig:pcie_8tp},~\ref{fig:a100_8tp}, and \ref{fig:h800_8tp} show performance results for ReduceScatter and AllGather overlapping. 
Given these evaluated sizes, 
\flux can deliver 1.20x to 3.25x speedups on A100 PCIe, 1.01x to 1.33x speedups on A100 NVLink, and 1.10x to 1.51x speedups on H800 NVLink over TransformerEngine. 
In terms of overlap efficiency, 
\flux can deliver 41\% to 57\% on A100 PCIe, 36\% to 96\% on A100 NVLink, and 37\% to 93\% on H800 NVLink, while TransformerEngine has -125\% to 36\% on A100 PCIe, -99\% to 74\% on A100 NVLink, and -40\% to 80\% on H800 NVLink. Note a negative overlap efficiency implies worse performance than the non-overlapping baseline.

Figure~\ref{fig:small_8tp} shows performance comparison for much smaller m sizes.
Given these evaluated sizes, 
\flux can deliver 1.45x to 3.21x speedups on A100 PCIe, 1.33x to 4.68x speedups on A100 NVLink, and a 0.95x slowdown to a 1.03 speedup on H800 NVLink over TransformerEngine. 
In terms of overlap efficiency, 
\flux delivers -2\% to 41\% on A100 PCIe, 14\% to 88\% on A100 NVLink, and -165\% to -82\% on H800 NVLink, 
while TransformerEngine has -213\% to -36\% on A100 PCIe, -325\% to -49\% on A100 NVLink, and -142\% to -93\% on H800 NVLink.

Figure~\ref{fig:tp16} shows performance comparison for 16-way tensor parallelism on 16-GPU cluster (8 GPUs per node and two nodes), with (m, n, k) as (8192, 49152, 12288) and (8192, 12288, 49152) in AllGather and ReduceScatter.
We only compare \flux with the PyTorch baseline, 
since TransformerEngine does not support multi-node overlapping. 
\flux can deliver up to 1.32x speedups and 18\% overlap efficiency on A100 PCIe, up to 1.57x speedups and 74\% overlap efficiency on A100 NVLink, 
and up to 1.55x speedups and 56\% overlap efficiency on H800 NVLink over PyTorch with fastest GEMM and NCCL.

\subsection{Model-level Performance Evaluation}
The evaluated models are GPT-3 175B and Llama-2 70B for both training and inference. 
In training, we use Megatron-LM core r0.4.0\footnote{commit 27cbe46}~\cite{megatron} for GPT-3 175B and 
Megatron-LLaMA~\cite{megatron-llama} for Llama-2 70B~\cite{touvron2023llama} on 128-GPU clusters with 2-way data, 8-way pipeline, and 8-way tensor parallelism. 
The entire training time, including gradient and optimizer phases, are reported.
In inference, we use vLLM 0.2.1~\cite{kwon2023efficient} for both models, with batch size as 8 and sequence length as 2048 in the prefill phase, and batch sizes as 64 or 512 in the decoding phase. 

\begin{figure*}[t]
\centering
\includegraphics[scale=0.4,bb=0 0 1250 350]{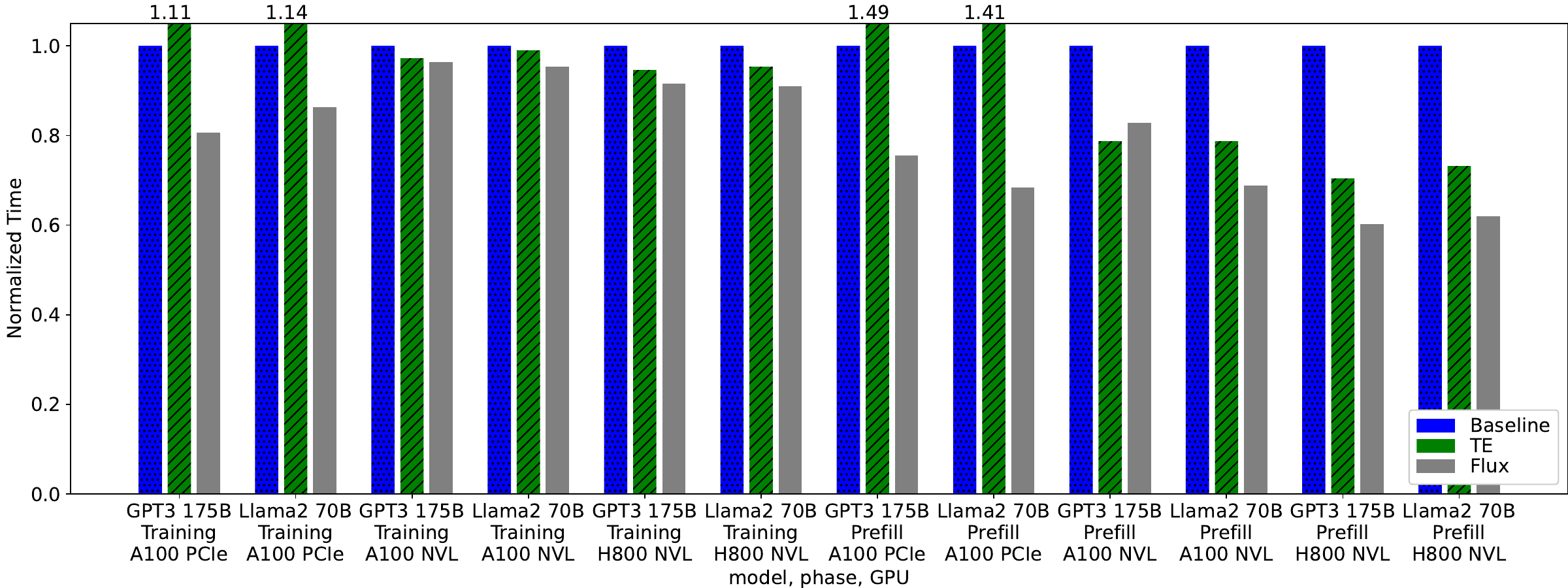}
\caption{Ene-to-End results for training and prefill of GPT-3 175B and Llama-2 70B on various GPU clusters.}
\label{fig:model_result}
\end{figure*}

\begin{figure*}[t]
\centering
\includegraphics[scale=0.4,bb=0 0 1250 350]{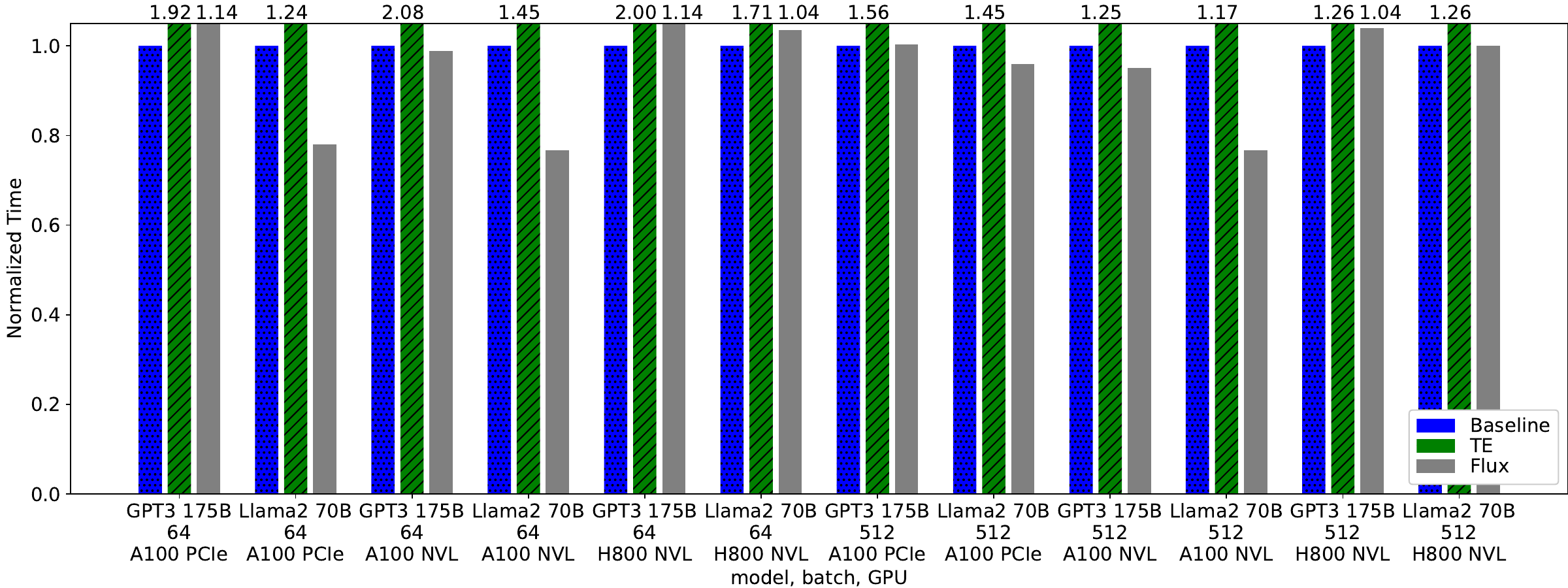}
\caption{End-to-End results for decoding of GPT-3 175B and Llama-2 70B on various 8-GPU clusters.}
\label{fig:decode_result}
\end{figure*}

Figure~\ref{fig:model_result} and \ref{fig:decode_result} show performance results for training, prefill, and decoding phases.
\flux can deliver up to 1.37x training, 2.06x prefill, and 1.69x decoding speedups on A100 PCIe, 
1.04x training, 1.14x prefill, and 2.10x decoding speedups on A100 NVLink, 
and 1.05x training, 1.18x prefill, and 1.76x decoding speedups on H800 NVLink over TransformerEngine, 
while \flux can deliver up to 1.24x training, 1.46x prefill, and 1.28x speedups on A100 PCIe, 
1.05x training, 1.45x prefill, and 1.30x decoding speedups on A100 NVLink, 
and 1.10x training, 1.66x prefill, and \textit{no} decoding speedups on H800 NVLink over Megatron-LM and vLLM baselines.

\section{Discussion}
\label{sec:dicuss}

\textbf{Small m sizes and decoding.}
When m is small (less than or equal to 1024), 
\flux outperforms TransformerEngine significantly by 1.03x to 4.68x speedups, except a 0.95x slowdown case of ReduceScatter with m as 64 on H800 NVLink. 
Figure~\ref{fig:small_8tp} echoes our earlier statement that the existing methods could underutilize GPU compute power by splitting GEMM kernels.
The effect becomes more apparent when the problem size getting small.
The figure also shows that \flux could perform worse than the non-overlapping baseline in a few extremely small m cases, while TransformerEngine performs even much worse in all small cases. 
That is mainly because when m is extremely small, the GEMM kernels typically have fewer warps, making latency hiding less efficient. 
Moreover, when m as 64 on H800 GPU, after 8-way tensor parallelism, 
\flux ReduceScatter further makes TMA instruction less efficient by reducing TMA store size to 8 along m, causing the only data point worse than TransformerEngine. 
Similar effects can be observed from the decoding results in Figure~\ref{fig:decode_result}, especially for the results with the batch size as 64. 
As the figure shown, \flux outperforms TransformerEngine by 1.21x to 2.10x speedups, but still has 5 cases slower than the non-overlapping vLLM baseline.
The results of the batch size 512 are better than ones of the batch size 64, aligning with the above mentioned reason.

\textbf{Overlap efficiency and speedups.}
The model-level performance is highly correlated to 2 factors, 1) tensor parallel communication portion shown in Figure~\ref{fig:comm_time}, and 2) overlap efficiency. 
When the communication portions are about 8\% to 11\% on A100 NVLink training, 
the overall speedups of \flux are only 1.04x to 1.05x over Megatron-LM and only 1.01x to 1.04x over TransformerEngine, 
regardless of \flux having 63\% average overlap efficiency on A100 NVLink cluster.
For the workloads with much higher communication portions, such as 40\% to 75\% on A100 PCIe training and prefill,
\flux can easily deliver 1.16x to 1.46x speedups over Megatron-LM, and 1.32x to 2.06x speedups over TransformerEngine.
As the results of the operation-level evaluation shown,  
\flux can deliver 40\%, 63\% and 72\% average overlap efficiency on A100 PCIe, A100 NVLink, and H800 NVLink clusters, respectively, while TransformerEngine only delivers -67\%, -61\%, and 20\% average overlap efficiency, respectively.
With that high overlap efficiency, \flux has a great potential to improve performance of tensor parallelism.

\textbf{High communication proportion.}
As just mentioned, the communication proportion is a key factor to take benefits from \flux. 
With high communication proportion on A100 PCIe and H800 NVLink clusters,
\flux outperforms TransformerEngine by noticeable speedups, up to 3.25x on A100 PCIe and 1.51x on H800 NVLink, in the operation level. 
These two clusters have high communication proportion for different reasons, A100 PCIe due to slow interconnects and H800 NVLink due to fast computation.
No matter the reason, \flux can still deliver high efficient communication overlapping, demonstrating its robustness, and the advantages fundamentally from the algorithm, optimization, and implementation sides. 
On the A100 PCIe cluster, we even observe that \flux sometimes can run faster than the non-overlapping communication only time, especially in large m sizes.
Although we tend to use the best communication library as the baseline in the evaluation from the best of our knowledge, 
the widely used, standard GPU communication library, NCCL, can still underperform in some problem sizes. 
This also demonstrates the performance of \flux and its auto-tuning mechanism adapting the interconnection well. 

\section{Related Work}
\label{sec:related}
Communication overlapping techniques are widely applied on distributed systems for various applications, using various techniques~\cite{danalis2005transformations, danalis2009mpi, marjanovic2010overlapping, gysi2016dcuda}. 
\flux uses kernel fusion and mainly targets at training and inference of large deep learning models.

With tensor parallelism for large deep learning models, 
prior works~\cite{jangda2022breaking, wang2023overlap, te} decompose coarse-grained operations into a sequence of medium-grained ones, and carefully schedule them to overlap communication with computation for TPUs and GPUs.
The work~\cite{pope2023efficiently} applies the same overlap technique with~\cite{wang2023overlap} to efficient inference. 
The work~\cite{jangda2022breaking} fuses communication with elementwise operations, overlapping with GEMM in a medium-grained fashion. 
In comparison, \flux applies much finer-grained decomposition for GPUs, fuses GEMM with communication, and supports both training and inference.  

Pipeline parallelism~\cite{huang2019gpipe, narayanan2019pipedream, narayanan2021efficient, yang2021pipemare, lamy2023breadth, li2023fold3d} is another common technique for large deep learning models. 
These work~\cite{lamy2023breadth, narayanan2021efficient, li2023fold3d} potentially can overlap some pipeline communication with gradient reduction, reducing pipeline bubbles.
Particularly, work~\cite{narayanan2021efficient, li2023fold3d} combine tensor parallelism, pipeline parallelism, and data parallelism together for large deep learning model training. 
\flux can be applied in addition to further improve performance.

Accelerating collective communication~\cite{rashidi2021enabling, jangda2022breaking, shah2023taccl, cowan2023mscclang, mscclpp} is another direction for improving network utilization. 
\ignore{
While ACE~\cite{rashidi2021enabling} proposed an accelerator alongside the network to handle the communication protocol, the other works~\cite{jangda2022breaking, shah2023taccl, cowan2023mscclang} synthesize efficient collective communication through compiler techniques.
Particularly, communication can be overlapped with computation with medium-grained decomposition~\cite{jangda2022breaking}, or overlapped with independent communication~\cite{shah2023taccl, cowan2023mscclang}.
}
\flux is a communication overlapping solution and can work with accelerated collective communication. 
Communication compression techniques~\cite{dryden2016communication, alistarh2017qsgd, wen2017terngrad, xu2021deepreduce, xu2021grace, fei2021efficient, wang2023zero++} for deep learning also improve network utilization by reducing communication sizes. 
\flux can be combined with the above methods.

ZeRO sharding techniques~\cite{rajbhandari2020zero, ren2021zero, rajbhandari2021zero, zhao2023pytorch, wang2023zero++} partition weights and/or gradients onto multiple devices with data parallelism, and perform AllGather operation before computations.
Those AllGather communications can be easily prefetched and overlapped with independent computation.
\flux can be applied to activations, weights and gradients, thus can be coupled with the above techniques.

\section{Conclusion}
Communication overlapping techniques are crucial for running large deep learning models with tensor parallelism. 
Conventional overlapping techniques perform poorly on GPUs.
The paper propose a novel technique, \flux, to resolve the issues.
By over-decomposing communication with corresponding computation, and fusing them into a single large kernel, the proposed technique can significantly reduce the exposed communication time and effectively improve system FLOPS utilization regardless of training or inference.

\section{Acknowledgements}
We would like to express our sincere gratitude to Zhekun Zhang, Dongyang Wang and Yawei Wen for their assistance and guidance throughout the development of FLUX. Their expertise and insights were instrumental in overcoming the challenges encountered during this project. We also thank all our colleagues who contributed their time and knowledge, providing support and encouragement.

\bibliographystyle{ieeetr}
\bibliography{main}

\begin{thebibliography}{10}

\bibitem{kaplan2020scaling}
J.~Kaplan, S.~McCandlish, T.~Henighan, T.~B. Brown, B.~Chess, R.~Child, S.~Gray, A.~Radford, J.~Wu, and D.~Amodei, ``Scaling laws for neural language models,'' 2020.

\bibitem{brown2020language}
T.~Brown, B.~Mann, N.~Ryder, M.~Subbiah, J.~D. Kaplan, P.~Dhariwal, A.~Neelakantan, P.~Shyam, G.~Sastry, A.~Askell, {\em et~al.}, ``Language models are few-shot learners,'' {\em Advances in neural information processing systems}, vol.~33, pp.~1877--1901, 2020.

\bibitem{chowdhery2022palm}
A.~Chowdhery, S.~Narang, J.~Devlin, M.~Bosma, G.~Mishra, A.~Roberts, P.~Barham, H.~W. Chung, C.~Sutton, S.~Gehrmann, P.~Schuh, K.~Shi, S.~Tsvyashchenko, J.~Maynez, A.~Rao, P.~Barnes, Y.~Tay, N.~Shazeer, V.~Prabhakaran, E.~Reif, N.~Du, B.~Hutchinson, R.~Pope, J.~Bradbury, J.~Austin, M.~Isard, G.~Gur-Ari, P.~Yin, T.~Duke, A.~Levskaya, S.~Ghemawat, S.~Dev, H.~Michalewski, X.~Garcia, V.~Misra, K.~Robinson, L.~Fedus, D.~Zhou, D.~Ippolito, D.~Luan, H.~Lim, B.~Zoph, A.~Spiridonov, R.~Sepassi, D.~Dohan, S.~Agrawal, M.~Omernick, A.~M. Dai, T.~S. Pillai, M.~Pellat, A.~Lewkowycz, E.~Moreira, R.~Child, O.~Polozov, K.~Lee, Z.~Zhou, X.~Wang, B.~Saeta, M.~Diaz, O.~Firat, M.~Catasta, J.~Wei, K.~Meier-Hellstern, D.~Eck, J.~Dean, S.~Petrov, and N.~Fiedel, ``Palm: Scaling language modeling with pathways,'' 2022.

\bibitem{smith2022using}
S.~Smith, M.~Patwary, B.~Norick, P.~LeGresley, S.~Rajbhandari, J.~Casper, Z.~Liu, S.~Prabhumoye, G.~Zerveas, V.~Korthikanti, E.~Zhang, R.~Child, R.~Y. Aminabadi, J.~Bernauer, X.~Song, M.~Shoeybi, Y.~He, M.~Houston, S.~Tiwary, and B.~Catanzaro, ``Using deepspeed and megatron to train megatron-turing nlg 530b, a large-scale generative language model,'' 2022.

\bibitem{ramesh2021zeroshot}
A.~Ramesh, M.~Pavlov, G.~Goh, S.~Gray, C.~Voss, A.~Radford, M.~Chen, and I.~Sutskever, ``Zero-shot text-to-image generation,'' 2021.

\bibitem{dehghani2023scaling}
M.~Dehghani, J.~Djolonga, B.~Mustafa, P.~Padlewski, J.~Heek, J.~Gilmer, A.~Steiner, M.~Caron, R.~Geirhos, I.~Alabdulmohsin, R.~Jenatton, L.~Beyer, M.~Tschannen, A.~Arnab, X.~Wang, C.~Riquelme, M.~Minderer, J.~Puigcerver, U.~Evci, M.~Kumar, S.~van Steenkiste, G.~F. Elsayed, A.~Mahendran, F.~Yu, A.~Oliver, F.~Huot, J.~Bastings, M.~P. Collier, A.~Gritsenko, V.~Birodkar, C.~Vasconcelos, Y.~Tay, T.~Mensink, A.~Kolesnikov, F.~Pavetić, D.~Tran, T.~Kipf, M.~Lučić, X.~Zhai, D.~Keysers, J.~Harmsen, and N.~Houlsby, ``Scaling vision transformers to 22 billion parameters,'' 2023.

\bibitem{radford2022robust}
A.~Radford, J.~W. Kim, T.~Xu, G.~Brockman, C.~McLeavey, and I.~Sutskever, ``Robust speech recognition via large-scale weak supervision,'' 2022.

\bibitem{Zhang_2022}
Y.~Zhang, D.~S. Park, W.~Han, J.~Qin, A.~Gulati, J.~Shor, A.~Jansen, Y.~Xu, Y.~Huang, S.~Wang, Z.~Zhou, B.~Li, M.~Ma, W.~Chan, J.~Yu, Y.~Wang, L.~Cao, K.~C. Sim, B.~Ramabhadran, T.~N. Sainath, F.~Beaufays, Z.~Chen, Q.~V. Le, C.-C. Chiu, R.~Pang, and Y.~Wu, ``Bigssl: Exploring the frontier of large-scale semi-supervised learning for automatic speech recognition,'' {\em IEEE Journal of Selected Topics in Signal Processing}, vol.~16, p.~1519–1532, Oct. 2022.

\bibitem{liu2023large}
X.~Liu, D.~McDuff, G.~Kovacs, I.~Galatzer-Levy, J.~Sunshine, J.~Zhan, M.-Z. Poh, S.~Liao, P.~D. Achille, and S.~Patel, ``Large language models are few-shot health learners,'' 2023.

\bibitem{wu2023bloomberggpt}
S.~Wu, O.~Irsoy, S.~Lu, V.~Dabravolski, M.~Dredze, S.~Gehrmann, P.~Kambadur, D.~Rosenberg, and G.~Mann, ``Bloomberggpt: A large language model for finance,'' 2023.

\bibitem{chen2021evaluating}
M.~Chen, J.~Tworek, H.~Jun, Q.~Yuan, H.~P. de~Oliveira~Pinto, J.~Kaplan, H.~Edwards, Y.~Burda, N.~Joseph, G.~Brockman, A.~Ray, R.~Puri, G.~Krueger, M.~Petrov, H.~Khlaaf, G.~Sastry, P.~Mishkin, B.~Chan, S.~Gray, N.~Ryder, M.~Pavlov, A.~Power, L.~Kaiser, M.~Bavarian, C.~Winter, P.~Tillet, F.~P. Such, D.~Cummings, M.~Plappert, F.~Chantzis, E.~Barnes, A.~Herbert-Voss, W.~H. Guss, A.~Nichol, A.~Paino, N.~Tezak, J.~Tang, I.~Babuschkin, S.~Balaji, S.~Jain, W.~Saunders, C.~Hesse, A.~N. Carr, J.~Leike, J.~Achiam, V.~Misra, E.~Morikawa, A.~Radford, M.~Knight, M.~Brundage, M.~Murati, K.~Mayer, P.~Welinder, B.~McGrew, D.~Amodei, S.~McCandlish, I.~Sutskever, and W.~Zaremba, ``Evaluating large language models trained on code,'' 2021.

\bibitem{jangda2022breaking}
A.~Jangda, J.~Huang, G.~Liu, A.~H.~N. Sabet, S.~Maleki, Y.~Miao, M.~Musuvathi, T.~Mytkowicz, and O.~Saarikivi, ``Breaking the computation and communication abstraction barrier in distributed machine learning workloads,'' in {\em Proceedings of the 27th ACM International Conference on Architectural Support for Programming Languages and Operating Systems}, pp.~402--416, 2022.

\bibitem{wang2023overlap}
S.~Wang, J.~Wei, A.~Sabne, A.~Davis, B.~Ilbeyi, B.~Hechtman, D.~Chen, K.~S. Murthy, M.~Maggioni, Q.~Zhang, {\em et~al.}, ``Overlap communication with dependent computation via decomposition in large deep learning models,'' in {\em Proceedings of the 28th ACM International Conference on Architectural Support for Programming Languages and Operating Systems, Volume 1}, pp.~93--106, 2023.

\bibitem{te}
NVIDIA, ``{TransformerEngine}.'' \url{https://github.com/NVIDIA/TransformerEngine}, 2022.

\bibitem{lamy2023breadth}
J.~Lamy-Poirier, ``Breadth-first pipeline parallelism,'' {\em Proceedings of Machine Learning and Systems}, vol.~5, 2023.

\bibitem{narayanan2021efficient}
D.~Narayanan, M.~Shoeybi, J.~Casper, P.~LeGresley, M.~Patwary, V.~Korthikanti, D.~Vainbrand, P.~Kashinkunti, J.~Bernauer, B.~Catanzaro, {\em et~al.}, ``Efficient large-scale language model training on gpu clusters using megatron-lm,'' in {\em Proceedings of the International Conference for High Performance Computing, Networking, Storage and Analysis}, pp.~1--15, 2021.

\bibitem{li2023fold3d}
F.~Li, S.~Zhao, Y.~Qing, X.~Chen, X.~Guan, S.~Wang, G.~Zhang, and H.~Cui, ``Fold3d: Rethinking and parallelizing computational and communicational tasks in the training of large dnn models,'' {\em IEEE Transactions on Parallel and Distributed Systems}, vol.~34, no.~5, pp.~1432--1449, 2023.

\bibitem{shah2023taccl}
A.~Shah, V.~Chidambaram, M.~Cowan, S.~Maleki, M.~Musuvathi, T.~Mytkowicz, J.~Nelson, O.~Saarikivi, and R.~Singh, ``$\{$TACCL$\}$: Guiding collective algorithm synthesis using communication sketches,'' in {\em 20th USENIX Symposium on Networked Systems Design and Implementation (NSDI 23)}, pp.~593--612, 2023.

\bibitem{cowan2023mscclang}
M.~Cowan, S.~Maleki, M.~Musuvathi, O.~Saarikivi, and Y.~Xiong, ``Mscclang: Microsoft collective communication language,'' in {\em Proceedings of the 28th ACM International Conference on Architectural Support for Programming Languages and Operating Systems, Volume 2}, pp.~502--514, 2023.

\bibitem{wang2023zero++}
G.~Wang, H.~Qin, S.~A. Jacobs, C.~Holmes, S.~Rajbhandari, O.~Ruwase, F.~Yan, L.~Yang, and Y.~He, ``Zero++: Extremely efficient collective communication for giant model training,'' {\em arXiv preprint arXiv:2306.10209}, 2023.

\bibitem{cutlass}
V.~Thakkar, P.~Ramani, C.~Cecka, A.~Shivam, H.~Lu, E.~Yan, J.~Kosaian, M.~Hoemmen, H.~Wu, A.~Kerr, M.~Nicely, D.~Merrill, D.~Blasig, F.~Qiao, P.~Majcher, P.~Springer, M.~Hohnerbach, J.~Wang, and M.~Gupta, ``{CUTLASS}.'' https://github.com/NVIDIA/cutlass, 2024.

\bibitem{megatron}
NVIDIA, ``{Megatron-LM}.'' \url{https://github.com/NVIDIA/Megatron-LM}, 2021.

\bibitem{kwon2023efficient}
W.~Kwon, Z.~Li, S.~Zhuang, Y.~Sheng, L.~Zheng, C.~H. Yu, J.~E. Gonzalez, H.~Zhang, and I.~Stoica, ``Efficient memory management for large language model serving with pagedattention,'' in {\em Proceedings of the ACM SIGOPS 29th Symposium on Operating Systems Principles}, 2023.

\bibitem{shoeybi2019megatron}
M.~Shoeybi, M.~Patwary, R.~Puri, P.~LeGresley, J.~Casper, and B.~Catanzaro, ``Megatron-lm: Training multi-billion parameter language models using model parallelism,'' {\em arXiv preprint arXiv:1909.08053}, 2019.

\bibitem{korthikanti2023reducing}
V.~A. Korthikanti, J.~Casper, S.~Lym, L.~McAfee, M.~Andersch, M.~Shoeybi, and B.~Catanzaro, ``Reducing activation recomputation in large transformer models,'' {\em Proceedings of Machine Learning and Systems}, vol.~5, 2023.

\bibitem{pope2023efficiently}
R.~Pope, S.~Douglas, A.~Chowdhery, J.~Devlin, J.~Bradbury, J.~Heek, K.~Xiao, S.~Agrawal, and J.~Dean, ``Efficiently scaling transformer inference,'' {\em Proceedings of Machine Learning and Systems}, vol.~5, 2023.

\bibitem{rajbhandari2020zero}
S.~Rajbhandari, J.~Rasley, O.~Ruwase, and Y.~He, ``Zero: Memory optimizations toward training trillion parameter models,'' in {\em SC20: International Conference for High Performance Computing, Networking, Storage and Analysis}, pp.~1--16, IEEE, 2020.

\bibitem{ren2021zero}
J.~Ren, S.~Rajbhandari, R.~Y. Aminabadi, O.~Ruwase, S.~Yang, M.~Zhang, D.~Li, and Y.~He, ``$\{$ZeRO-Offload$\}$: Democratizing $\{$Billion-Scale$\}$ model training,'' in {\em 2021 USENIX Annual Technical Conference (USENIX ATC 21)}, pp.~551--564, 2021.

\bibitem{rajbhandari2021zero}
S.~Rajbhandari, O.~Ruwase, J.~Rasley, S.~Smith, and Y.~He, ``Zero-infinity: Breaking the gpu memory wall for extreme scale deep learning,'' in {\em SC21: International Conference for High Performance Computing, Networking, Storage and Analysis}, pp.~1--15, IEEE, 2021.

\bibitem{zhao2023pytorch}
Y.~Zhao, A.~Gu, R.~Varma, L.~Luo, C.-C. Huang, M.~Xu, L.~Wright, H.~Shojanazeri, M.~Ott, S.~Shleifer, {\em et~al.}, ``Pytorch fsdp: experiences on scaling fully sharded data parallel,'' {\em arXiv preprint arXiv:2304.11277}, 2023.

\bibitem{jiang2024megascale}
Z.~Jiang, H.~Lin, Y.~Zhong, Q.~Huang, Y.~Chen, Z.~Zhang, Y.~Peng, X.~Li, C.~Xie, S.~Nong, {\em et~al.}, ``Megascale: Scaling large language model training to more than 10,000 gpus,'' in {\em 21st USENIX Symposium on Networked Systems Design and Implementation (NSDI 24')}, 2024.

\bibitem{nccl}
{NVIDIA}, ``{NCCL}.'' https://github.com/NVIDIA/nccl, 2016.

\bibitem{nvshmem}
NVIDIA, ``{NVSHMEM}.'' \url{https://developer.nvidia.com/nvshmem}, 2020.

\bibitem{evt}
Z.~Chen, A.~Kerr, R.~Cai, J.~Kosaian, H.~Wu, Y.~Ding, and Y.~Xie, ``{EVT}: Accelerating deep learning training with epilogue visitor tree,'' in {\em Proceedings of the 29th ACM International Conference on Architectural Support for Programming Languages and Operating Systems}, 2024 (in press).

\bibitem{osama2023stream}
M.~Osama, D.~Merrill, C.~Cecka, M.~Garland, and J.~D. Owens, ``Stream-k: Work-centric parallel decomposition for dense matrix-matrix multiplication on the gpu,'' in {\em Proceedings of the 28th ACM SIGPLAN Annual Symposium on Principles and Practice of Parallel Programming}, pp.~429--431, 2023.

\bibitem{megatron-llama}
Alibaba, ``{Megatron-LLaMA}.'' \url{https://github.com/alibaba/Megatron-LLaMA}, 2023.

\bibitem{touvron2023llama}
H.~Touvron, L.~Martin, K.~Stone, P.~Albert, A.~Almahairi, Y.~Babaei, N.~Bashlykov, S.~Batra, P.~Bhargava, S.~Bhosale, {\em et~al.}, ``Llama 2: Open foundation and fine-tuned chat models,'' {\em arXiv preprint arXiv:2307.09288}, 2023.

\bibitem{danalis2005transformations}
A.~Danalis, K.-Y. Kim, L.~Pollock, and M.~Swany, ``Transformations to parallel codes for communication-computation overlap,'' in {\em SC'05: Proceedings of the 2005 ACM/IEEE conference on Supercomputing}, pp.~58--58, IEEE, 2005.

\bibitem{danalis2009mpi}
A.~Danalis, L.~Pollock, M.~Swany, and J.~Cavazos, ``Mpi-aware compiler optimizations for improving communication-computation overlap,'' in {\em Proceedings of the 23rd international conference on Supercomputing}, pp.~316--325, 2009.

\bibitem{marjanovic2010overlapping}
V.~Marjanovi{\'c}, J.~Labarta, E.~Ayguad{\'e}, and M.~Valero, ``Overlapping communication and computation by using a hybrid mpi/smpss approach,'' in {\em Proceedings of the 24th acm International Conference on Supercomputing}, pp.~5--16, 2010.

\bibitem{gysi2016dcuda}
T.~Gysi, J.~B{\"a}r, and T.~Hoefler, ``dcuda: hardware supported overlap of computation and communication,'' in {\em SC'16: Proceedings of the International Conference for High Performance Computing, Networking, Storage and Analysis}, pp.~609--620, IEEE, 2016.

\bibitem{huang2019gpipe}
Y.~Huang, Y.~Cheng, A.~Bapna, O.~Firat, D.~Chen, M.~Chen, H.~Lee, J.~Ngiam, Q.~V. Le, Y.~Wu, {\em et~al.}, ``Gpipe: Efficient training of giant neural networks using pipeline parallelism,'' {\em Advances in neural information processing systems}, vol.~32, 2019.

\bibitem{narayanan2019pipedream}
D.~Narayanan, A.~Harlap, A.~Phanishayee, V.~Seshadri, N.~R. Devanur, G.~R. Ganger, P.~B. Gibbons, and M.~Zaharia, ``Pipedream: Generalized pipeline parallelism for dnn training,'' in {\em Proceedings of the 27th ACM Symposium on Operating Systems Principles}, pp.~1--15, 2019.

\bibitem{yang2021pipemare}
B.~Yang, J.~Zhang, J.~Li, C.~R{\'e}, C.~Aberger, and C.~De~Sa, ``Pipemare: Asynchronous pipeline parallel dnn training,'' {\em Proceedings of Machine Learning and Systems}, vol.~3, pp.~269--296, 2021.

\bibitem{rashidi2021enabling}
S.~Rashidi, M.~Denton, S.~Sridharan, S.~Srinivasan, A.~Suresh, J.~Nie, and T.~Krishna, ``Enabling compute-communication overlap in distributed deep learning training platforms,'' in {\em 2021 ACM/IEEE 48th Annual International Symposium on Computer Architecture (ISCA)}, pp.~540--553, IEEE, 2021.

\bibitem{mscclpp}
Microsoft, ``{MSCCL++}.'' \url{https://github.com/microsoft/mscclpp}, 2023.

\bibitem{dryden2016communication}
N.~Dryden, T.~Moon, S.~A. Jacobs, and B.~Van~Essen, ``Communication quantization for data-parallel training of deep neural networks,'' in {\em 2016 2nd Workshop on Machine Learning in HPC Environments (MLHPC)}, pp.~1--8, IEEE, 2016.

\bibitem{alistarh2017qsgd}
D.~Alistarh, D.~Grubic, J.~Li, R.~Tomioka, and M.~Vojnovic, ``Qsgd: Communication-efficient sgd via gradient quantization and encoding,'' {\em Advances in neural information processing systems}, vol.~30, 2017.

\bibitem{wen2017terngrad}
W.~Wen, C.~Xu, F.~Yan, C.~Wu, Y.~Wang, Y.~Chen, and H.~Li, ``Terngrad: Ternary gradients to reduce communication in distributed deep learning,'' {\em Advances in neural information processing systems}, vol.~30, 2017.

\bibitem{xu2021deepreduce}
H.~Xu, K.~Kostopoulou, A.~Dutta, X.~Li, A.~Ntoulas, and P.~Kalnis, ``Deepreduce: A sparse-tensor communication framework for federated deep learning,'' {\em Advances in Neural Information Processing Systems}, vol.~34, pp.~21150--21163, 2021.

\bibitem{xu2021grace}
H.~Xu, C.-Y. Ho, A.~M. Abdelmoniem, A.~Dutta, E.~H. Bergou, K.~Karatsenidis, M.~Canini, and P.~Kalnis, ``Grace: A compressed communication framework for distributed machine learning,'' in {\em 2021 IEEE 41st international conference on distributed computing systems (ICDCS)}, pp.~561--572, IEEE, 2021.

\bibitem{fei2021efficient}
J.~Fei, C.-Y. Ho, A.~N. Sahu, M.~Canini, and A.~Sapio, ``Efficient sparse collective communication and its application to accelerate distributed deep learning,'' in {\em Proceedings of the 2021 ACM SIGCOMM 2021 Conference}, pp.~676--691, 2021.

\end{thebibliography}

\end{document}